\documentclass[letterpaper]{article} 
\usepackage{aaai24}  
\usepackage{times}  
\usepackage{helvet}  
\usepackage{courier}  
\usepackage[hyphens]{url}  
\usepackage{graphicx} 
\usepackage{natbib}  
\usepackage{caption} 
\usepackage{algorithm}

\usepackage{multirow}
\usepackage{rotating}
\usepackage{makecell}
\usepackage{subfig}
\usepackage{algorithmicx}
\usepackage{algpseudocode}
\usepackage{amssymb}
\usepackage{amsmath}
\usepackage{arydshln}
\algnewcommand{\algorithmicinput}{\textbf{Input:}}
\algnewcommand{\algorithmicoutput}{\textbf{Output:}}
\algnewcommand\Input{\item[\algorithmicinput]}%
\algnewcommand\Output{\item[\algorithmicoutput]}%
\newcommand{\argmin}{\operatornamewithlimits{argmin}}
\newcommand{\argmax}{\operatornamewithlimits{argmax}}
\newcommand{\OursFL}{TurboSVM-FL }
\newcommand{\OursFLNoSpace}{TurboSVM-FL}

\usepackage{newfloat}
\usepackage{listings}
\DeclareCaptionStyle{ruled}{labelfont=normalfont,labelsep=colon,strut=off} 
\floatstyle{ruled}
\newfloat{listing}{tb}{lst}{}
\floatname{listing}{Listing}
\title{TurboSVM-FL: Boosting Federated Learning through SVM Aggregation for Lazy Clients}
\author{
    Mengdi Wang\equalcontrib,
    Anna Bodonhelyi\equalcontrib,
    Efe Bozkir,
    Enkelejda Kasneci
}
\affiliations{
    
    Chair for Human-Centered Technologies for Learning, Technical University of Munich, Munich, Bavaria, Germany \\

    \{mengdi.wang, anna.bodonhelyi, efe.bozkir, enkelejda.kasneci\}@tum.de \\ %
}

\usepackage{bibentry}

\begin{document}

\maketitle

\begin{abstract}
\fontsize{9}{10}\selectfont
Federated learning is a distributed collaborative machine learning paradigm that has gained strong momentum in recent years. In federated learning, a central server periodically coordinates models with clients and aggregates the models trained locally by clients without necessitating access to local data. Despite its potential, the implementation of federated learning continues to encounter several challenges, predominantly the slow convergence that is largely due to data heterogeneity. The slow convergence becomes particularly problematic in cross-device federated learning scenarios where clients may be strongly limited by computing power and storage space, and hence counteracting methods that induce additional computation or memory cost on the client side such as auxiliary objective terms and larger training iterations can be impractical. In this paper, we propose a novel federated aggregation strategy, \textbf{~\OursFLNoSpace}, that poses no additional computation burden on the client side and can significantly accelerate convergence for federated classification task, especially when clients are ``lazy" and train their models solely for few epochs for next global aggregation.~\OursFL extensively utilizes support vector machine to conduct selective aggregation and max-margin spread-out regularization on class embeddings. We evaluate~\OursFL on multiple datasets including FEMNIST, CelebA, and Shakespeare using user-independent validation with non-iid data distribution. Our results show that~\OursFL can significantly outperform existing popular algorithms on convergence rate and reduce communication rounds while delivering better test metrics including accuracy, F1 score, and MCC. 
\end{abstract}

\section{Introduction}\label{chap:introduction}
With the increasing importance of data privacy, a giant stride in distributed machine learning has been observed in recent years. As one promising distributed machine learning paradigm, federated learning (FL) has been growing at an astounding rate after its introduction~\cite{mcmahan2017communication}. In the common FL settings, data is distributed over numerous end clients, while the central server possesses no data by itself. After the server initiates a model and sends the model to clients, each client trains the model locally using its own data. The server periodically aggregates the locally trained models and synchronizes local models of clients with the latest aggregated one. With such a process, FL provides a primary privacy guarantee to a large extent since the server does not require data sharing and is hence preferred in many privacy-preserving scenarios where sensitive data is utilized. 

Based on the characteristics of participating entities, FL can be further categorized into cross-silo FL and cross-device FL~\cite{kairouz2021advances}. In cross-silo FL, the target clients are often large-scale institutions such as hospitals, data centers, educational organizations, and high-tech companies. Such stakeholders commonly possess decent resource for computing, storage, and internet connection, while the number of attending institutions is relatively low. Therefore, the probability that each client takes part in all aggregation rounds is high. In contrast to cross-silo FL, cross-device FL focuses more on training on end-user devices like smartphones and personal computers using user data. The scale of participating clients in cross-device FL can be fairly large, while each client may be strongly limited by its computing power and connectivity. As a consequence, only a (small) portion of clients could share their models during a global aggregation. An additional critical aspect of cross-device FL is that each client might only contain data collected from a single user, which exacerbates data heterogeneity across clients.
\begin{figure*}[t]
\centering
\includegraphics[width=17.5cm, keepaspectratio]{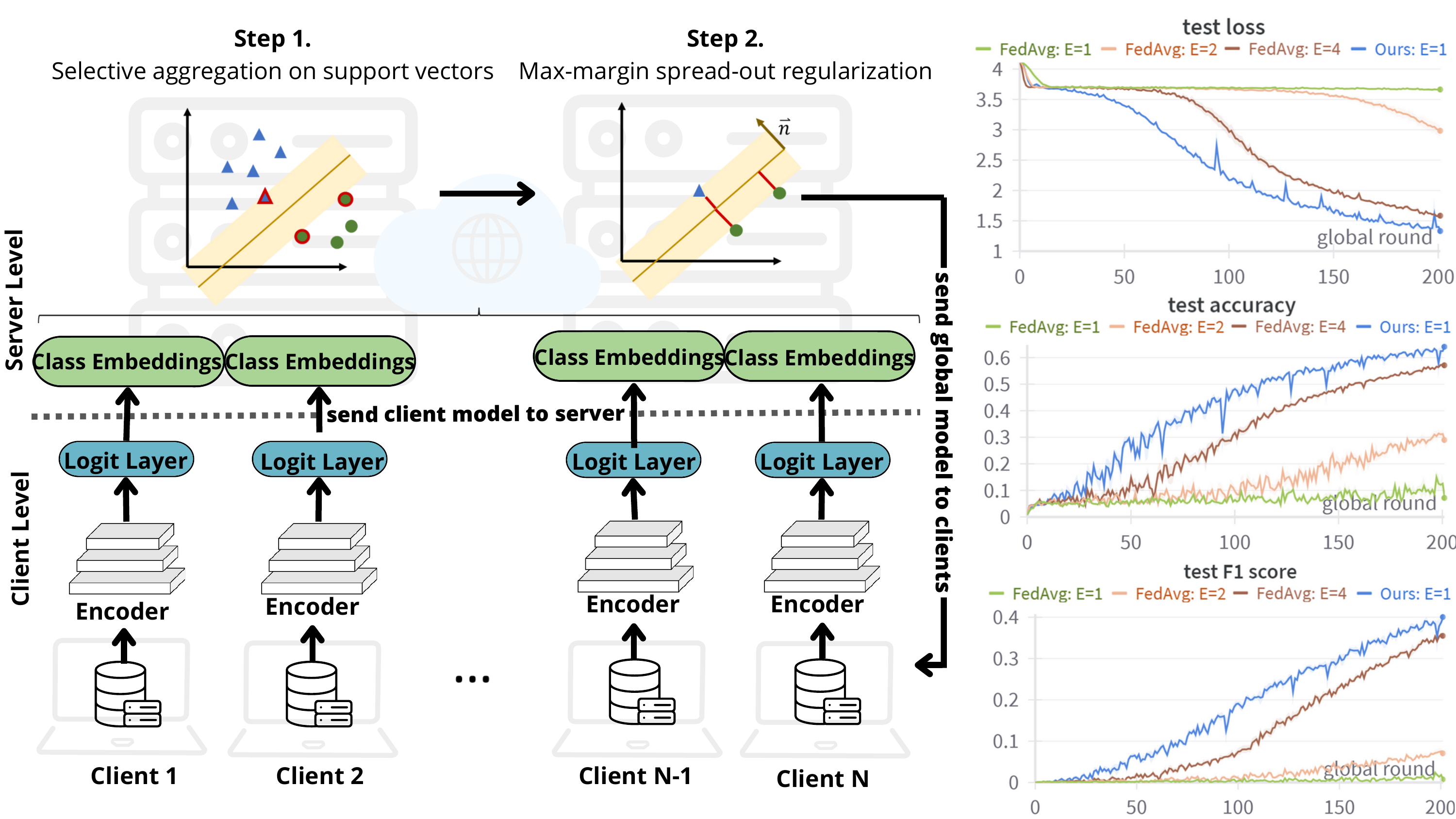}
\caption{Left: pipeline of~\OursFLNoSpace. Right: test performance of~\OursFL against FedAvg. $E$ indicates the number of client local training epochs. The results were obtained on FEMNIST dataset using a suboptimal client learning rate.
\label{fig:teaser}
}
\end{figure*}

Despite recent advancements in FL, its implementation in practice is still facing some challenges. Among these, the slow convergence is a primary concern: a significantly greater number of aggregation rounds are often needed to reach convergence compared to non-FL setups. Several factors contribute to this inefficiency according to~\cite{wu2023faster}, such as client drift caused by data heterogeneity~\cite{karimireddy2020scaffold}, lack of adaptive optimization~\cite{reddi2020adaptive}, and the increase in model complexity and data size. One trivial solution proposed in~\cite{mcmahan2017communication} is to increase client computation load by a larger number of local training iterations. Although this solution vastly speeds up the convergence, it multiplies the computation load on the client side, which can be problematic when clients are constrained by computing power, like in cross-device FL case. Other existing solutions mainly target at either client drift or adaptive optimization. The former often requires to optimize additional objective functions on the client side, which in turn also increases the client computation load and even the need for storage, while the latter can be particularly hard to tune because it is often necessary to decide the choice of optimizers and learning rates jointly between server and clients. There also exist solutions that require additional data on the server side, which may increase the risk of privacy leaks if not handled properly.

In this paper, we focus on embedding-based neural network classifiers, which means the input samples are encoded into the same space as class representations and their similarity determines the probability of the sample belonging to the class~\cite{yu2020federated}. We propose a novel federated aggregation strategy for classification tasks called~\OursFLNoSpace.~\OursFL induces no additional computation cost or storage consumption on the client side compared to the vanilla FL algorithm, and shows great potential in reducing communication rounds, especially when clients are ``lazy" and train their local models for very few iterations.~\OursFL extensively exploits the property of support vector machine (SVM) and consists of three steps. Firstly, \OursFL reformalizes classification while treating client models in a model-as-sample way, and fits SVMs using latent class embeddings. Then, it conducts selective aggregation on latent class features that form support vectors. Lastly,~\OursFL applies max-margin spread-out regularization on aggregated class representations upon SVM hyperplanes. By adopting adaptive methods in max-margin spread-out regularization,~\OursFL can also benefit from adaptive optimization. The main contributions of this work can be summarized as follows:
\begin{itemize}
    \item We introduce a novel perspective to interpret classification in FL setting and a model-as-sample strategy, which lay the foundation for further FL improvements such as selective aggregation and outlier detection.
    
    \item We propose a novel federated aggregation algorithm named~\OursFL that vastly boosts convergence for federated classification tasks using deep neural networks. The proposed method extensively exploits support vector machine (SVM) to conduct selective model aggregations and max-margin spread-out regularization.
    
    \item We conduct experiments on various benchmarks in user-independent validation and show the potential of~\OursFL in reducing communication cost. The benchmarks contain three different tasks covering image classification and natural language processing with non-iid data distribution over clients.
\end{itemize}

\section{Related Work}\label{chap:relwork}

\subsection{Federated Learning}
The concept of FL was originally introduced in~\cite{mcmahan2017communication}. Unlike centralized learning where the goal is to fit an optimal model on a collection of data, FL aims to train a model that delivers superior performance across data subsets. In the remaining of this work, we narrow down our focus to federated classification tasks. In a federated classification task with $K$ classes and $N$ clients, denote the local dataset of each client as $D_1,..., D_N$ with $D_n = \{(x, y)\}, n \in [N]$, where $(x, y) \in \mathbb{R}^P \times [K]$ is a sample point of class $y \in [K]$. Let $D_G = \biguplus_{n=1}^{N} D_n$ with $|D_G| = \sum_{n=1}^{N} |D_n|$ describe the collection of local datasets and $\ell(x, y, \theta)$ be the objective function measured on the sample $(x, y)$ with model $\theta$. Then, the goal of centralized learning is to find an optimal model $\theta^*$ that satisfies:
\begin{equation}
    \theta^* = \argmin_\theta \mathbb{E}_{(x,y) \sim D_G} [\ell(x, y, \theta)]
\end{equation}
In contrast, FL aims to fit a model that performs optimally across clients:
\begin{equation}
    \theta^* = \argmin_\theta \sum_{n=1}^{N} \frac{|D_n|}{|D_G|} \mathbb{E}_{(x,y) \sim D_n} [\ell(x, y, \theta)]
\end{equation}

The typical workflow of FL can be broken down into three steps. First, a server initializes a model and broadcasts this model to all clients. Then, each client trains the received model on its own dataset for $E$ epochs and sends the trained model back to the server. In the next step, the server aggregates locally trained models into a new global model and synchronizes clients with the latest global model. The last two steps are repeated for multiple rounds until convergence. The first federated aggregation algorithm, FedAvg, was introduced in~\cite{mcmahan2017communication} and applies weighted average over client models:
\begin{equation}
    \theta_G = \sum_{n=1}^{N} \frac{|D_n|}{|D_G|} \theta_n, \theta_n = \argmin_\theta  \mathbb{E}_{(x,y) \sim D_n} [\ell(x, y, \theta)]
    \label{eq:fedavg}
\end{equation}
where $\theta_G$ and $\theta_n$ denote the aggregated global model and the local model of $n$-th client, respectively.

One of those challenges that FL is facing is the large amount of aggregation rounds needed to approach convergence. While increasing local training iterations $E$ can significantly advance convergence, it also vastly increases the computation load on the client side, which can be extremely problematic in cross-device FL. Many follow-up works aim to speed up FL convergence, and they can be mainly categorized into two groups. The first group endeavor to address client drift~\cite{karimireddy2020scaffold} caused by data heterogeneity, while the other group attempt to benefit FL with adaptive learning methods, which we describe as follows.

\subsection{Client Drift in Federated Learning}
Client drift~\cite{karimireddy2020scaffold} describes the phenomenon that client local models approach individual local optima rather than global optima and their average is drifted away from global optima as well, which is caused by data heterogeneity across client local datasets and can dramatically impact convergence behavior. Various recent works attempt to solve client drift on the client side. SCAFFOLD~\cite{karimireddy2020scaffold} introduces a control variate term to stabilize gradient update. FedProx~\cite{li2020federated} proposes an additional loss term based on L2 distance between global model $\theta_G$ and client model $\theta_n$ during local training. MOON~\cite{li2021model} and FedProc~\cite{mu2023fedproc} suggest the use of contrastive learning to combat data heterogeneity. The former introduces an objective based on latent features extracted respectively by the global model, current client model, and previous client model, while the latter penalizes the dissimilarity between latent features and class representations. A common drawback of the aforementioned methods lies in that they increase either computation burden or memory consumption or even both on the client side, which can be quite a challenge for end-user devices like smartphones and tablets. 

There also exists works that aim to solve client drift on the server side. For instance, FedAwS~\cite{yu2020federated} addresses an extreme data distribution case where clients may have only data from a single class. FedAwS utilizes spread-out regularizer~\cite{zhang2017learning} and raises a penalty term based on cosine similarities among class embeddings on the server side.

\subsection{Adaptive Federated Learning}
In centralized learning, advanced adaptive and momentum-based optimization techniques such as AdaGrad~\cite{duchi2011adaptive}, Adam~\cite{kingma2014adam}, and Yogi~\cite{zaheer2018adaptive} have shown great success in convergence acceleration. In contrast, in vanilla FedAvg, client models are trained with stochastic gradient descent~\cite{robbins1951stochastic, kiefer1952stochastic} and server aggregation is (weighted) averaging. Numerous works have been devoted to benefiting FL with advanced server-side optimizers. As a forerunner in this field,~\cite{reddi2020adaptive} proposed a family of adaptive aggregation methods called FedOpt. Different from weighted average in Equation~\ref{eq:fedavg}, FedOpt computes pseudo-gradient~\cite{chen2016scalable, nichol2018first} from client models and updates the global model with a chosen optimizer:
\begin{gather}
    \Delta \gets \sum_{n=1}^{N} \frac{|D_n|}{|D_G|}\theta_n - \theta_G \\
    \theta_G \gets \text{server\_optimizer}(\theta_G, -\Delta, \eta)
\end{gather}
where $\eta$ indicates the learning rate. Depending on the choice of optimizer, FedOpt can be derivated into multiple variants. For instance, in~\cite{reddi2020adaptive}, the researchers introduced FedAdaGrad, FedAdam, and FedYogi with their names indicating the choice of optimization technique. FedAMS~\cite{wang2022communication} suggests the use of AMSGrad~\cite{reddi2019convergence} optimizer on the server side, which is an improved version of Adam. According to~\cite{wang2021field}, it can be hard to tune FedOpt-family methods due to the additional implementation of optimizer on the server side, and it is often necessary to search for optimal learning rates jointly for client optimizer and server optimizer. 

Compared to server-level adaptive learning, adaptive optimization on the client side is less studied. Client adaptivity poses its own challenges, particularly due to the potential for the states of client optimizers to significantly diverge from each other as a result of data heterogeneity. To address this challenge,~\cite{wang2021local} proposes to reset client optimizer status in each global round, while LocalAMSGrad~\cite{chen2020toward} and FAFED~\cite{wu2023faster} suggest the sharing and aggregation of client optimizer states similarly to client models. These methods can be meaningless if clients are limited by computation resource and can only train their local models for few epochs.

\subsection{Support Vector Machine}
Support vector machine (SVM)~\cite{cortes1995support} is a widely-used and robust supervised learning model that can be used for both regression and classification tasks. Unlike traditional linear classifiers, where the decision boundary is a linear combination of all data points, the separating hyperplane of SVM is a combination of selected samples, which are also called support vectors and lie the closest to the decision boundary. While common classifiers minimize solely the classification objective, SVM struggles to control the trade-off between discriminative error minimization and margin maximization while allowing some misclassifications. The margin refers to the distance between the support vectors of different classes and the decision boundary. The primal problem of SVM can be formalized as:
\begin{equation}
    \argmin_{w, \zeta_1, ..., \zeta_m} \frac{1}{2} ||w||^2 + \lambda \sum_{i=1}^{m} \zeta_i, \text{s.t. } y_i w^\tau x_i \geq 1 - \zeta_i, \zeta_i \geq 0
    \label{eq:svm}
\end{equation}
where $\zeta_i$ defines the distance of a misclassified sample to its correct margin plane. The coefficient $\lambda$ controls the magnitude of regularization. A smaller $\lambda$ prioritizes larger margins and may result in a greater number of support vectors.

It is important to note that there are several prior works that integrate FL and SVM, such as~\cite{bakopoulou2021fedpacket, navia2022budget, bemani2022aggregation}. Our approach is distinctly different from them in the sense that our algorithm leverages SVM to improve global aggregation and offers a solution to the problem ``how to FL''. In contrast, in previous works, SVM serves as the core model to be trained in FL and thus addresses the question of ``what to FL".

\section{Methodology}
In this paper, we propose a novel federated aggregation algorithm for classification task called~\OursFL that is able to boost convergence significantly.~\OursFL sophisticatedly leverages SVM to conduct selective aggregation and max-margin spread-out regularization. By adopting an adaptive optimizer,~\OursFL can also benefit from adaptivity. Compared to vanilla FedAvg~\cite{mcmahan2017communication},~\OursFL requires no additional computation, storage, or communication cost on the client side. A pseudocode for~\OursFL is given in Algorithm~\ref{alg:ours_all}, and a graphical illustration is depicted Figure~\ref{fig:teaser}.

In the following, we present our algorithm in detail, starting by reformalizing the classification task as ``finding nearest neighbor". We reduce our discussion to embedding-based deep learning networks and ignore the logit activation, which means the model $\theta$ can be divided into two parts: $g$ and $W$, with $f_{\theta}(x) = Wg(x)$, where $g: \mathbb{R}^P \rightarrow \mathbb{R}^d$ is the feature encoder that maps input $x \in \mathbb{R}^P$ to a latent representation in $\mathbb{R}^d$ and $W$ is the last projection layer containing class embeddings. In a classification task with $K$ classes, $W$ will be of shape $R^{K \times d}$ and is also called logit layer. Then, the class inference $\hat y$ of a sample $(x, y)$ is indifferent from finding the nearest neighbor to $g(x)$ in $w_1, ..., w_K$ with $w_k \in \mathbb{R}^d$ being the $k$-th row in $W$ indicating the class embedding of class $k$. Implicitly, the metric used to measure distance is vector inner product, and the choice of nearest neighbor is regardless of activation function and loss function. For simplicity, we ignore bias terms, and the class inference can then be represented as:
\begin{gather}
    \hat y = \argmax_{k \in [K]} \text{sim} (w_k, g(x)), \text{sim} (w_k, g(x)) = w_k ^ \tau \cdot g(x)
\end{gather}
Hence, training the last layer in classification can be regarded as encouraging the correct class embedding to approach instance embedding while discouraging all other class embeddings to be close.

Next,~\OursFL treats client models as sample points for a secondary classification task at higher level, and fits SVM using these samples. The SVM is constructed as a multi-class classification among $K$ classes, and the SVM training samples are exactly the collected class embeddings, i.e., $\{(w_k^n, k) | k \in [K], n \in [N]\}$. In other words, for each class $k$, there are $N$ sample points $\{(w_k^1, k), ..., (w_k^N, k)\}$, and each sample point is the $k$-th row of the weight matrix of the logit layer from a client model.

In vanilla FL, the class embeddings in the logit layer of the global model are obtained by averaging client models, in other words, $w_k^G = \sum_{n=1}^{N} \frac{|D_n|}{|D_G|} w_k^n$. Due to data heterogeneity among clients, the class embeddings of some clients can be drifted away from global optima and hence seriously disturb the aggregation.~\OursFL addresses this problem with the help of support vectors during global update. SVM aims at a margin-maximization decision boundary that is a linear combination of selected samples, which are also called support vectors. The support vectors can be regarded as the most informative samples of each class and function similarly to contrastive anchors. In other words, fitting SVM is to some extent equivalent to selective aggregation over samples.~\OursFL brings this property to federated aggregation by averaging only class embeddings that form support vectors, as depicted in Algorithm~\ref{alg:ours_1}.
\begin{algorithm}[b]
\caption{~\OursFL part 1: selective aggregation}\label{alg:ours_1}
\begin{algorithmic}
\State \textbf{Input:} fitted SVM, sizes of local datasets $|D_1|,...,|D_N|$.

\hfill
\For{$k \in [K]$}
\State retrieve support vectors $\{v_k^m\}$ for class $k$ from SVM
\State $w_k^G \gets \frac{\sum_m|D_m|v_k^m}{\sum_m|D_m|}$ \Comment{$m$: index to client model}
\EndFor
\State \textbf{return} global class embeddings $w_1^G,..., w_K^G$
\end{algorithmic}
\end{algorithm}

\begin{algorithm}[ht]
\caption{~\OursFL part 2: max-margin spread-out regularization}\label{alg:ours_2}
\begin{algorithmic}
\State \textbf{Input:} fitted SVM, global class embeddings $w_k^G, k \in [K]$, server learning rate $\eta_G$.

\hfill
\State $\ell_{sp} \gets 0$
\For{$k \in [K]$}
\For{$k' \in [K]$ with $k' > k$}
\State retrieve hyperplane $h_{k, k'}$ from SVM
\State $\ell_{sp} \gets \ell_{sp} + exp(-\frac{({w_k^G} ^ \tau \cdot h_{k, k'} - {w_{k'}^G} ^ \tau \cdot h_{k, k'})^2}{2||h_{k, k'}||^2})$
\EndFor
\EndFor
\For{$k \in [K]$}
\State $w_k^G \gets \text{server\_optimizer}(w_k^G, -\nabla_{w_k^G}\ell_{sp}, \eta_G)$
\EndFor
\State \textbf{return} global class embeddings $w_1^G,..., w_K^G$
\end{algorithmic}
\end{algorithm}

Moreover,~\OursFL employs spread-out regularization across projected global class embeddings to maintain the margin-maximization property. This is crucial for two reasons: first, although support vectors are the most informative data points, they are close to decision boundary and can be misclassified; second, the weights used during FL aggregation may differ from the coefficients assigned to support vectors during SVM fitting, which could undermine the SVM property. Spread-out regularization like in~\cite{yu2020federated} offers the potential to distinguish class embeddings. Nevertheless, we propose that omnidirectional regularization is not the most efficient method. Instead, we leverage once again the SVM property, namely we project the aggregated embeddings back onto the SVM decision boundaries, and penalize the similarities among projected embeddings:
\begin{equation}
    \ell_{sp}(w_1, ..., w_K) = \sum_{k \in [K]} \sum_{k' \neq k} \text{sim} (\frac{w_k ^ \tau \cdot h_{k, k'}}{||h_{k, k'}||}, \frac{w_{k'} ^ \tau \cdot h_{k, k'}}{||h_{k, k'}||})
\end{equation}
where $h_{k, k'}$ is the normal of the separating hyperplane for classes $k$ and $k'$ retrieved from fitted SVM. 

In~\cite{yu2020federated}, the authors proved that the classification error can be upper-bounded by the separation of class embeddings. We extend their analysis for~\OursFL by showing that by applying selective aggregation and max-margin spread-out regularization,~\OursFL effectively enlarges the difference between projected logits. For simplicity, we narrow down to binary classification and denote the distance relaxation terms for embeddings of each class as $\zeta_n^+$ and $\zeta_n^-, n \in [N]$, respectively. Let $h$ be the decision boundary of the fitted SVM. Then, under further simplification that all class embeddings serve as support vectors and all clients have same amount of samples, given a new positive sample $x^*$, the difference between the positive and negative logits when projected on $h$ can be bounded as follows:
\begin{align}    
\begin{split}
 & \text{proj(logit$^+(x^*)$, $h$)} - \text{proj(logit$^-(x^*)$, $h$)} \\
\geq& \frac{[2N - \sum_{n=1}^{N}(\zeta_n^+ + \zeta_n^-)](1 - \zeta^*)}{N ||h||^2}
\end{split}
\end{align}
where $\zeta^*$ is the SVM relaxation term for $x^*$. By averaging support vectors and applying max-margin spread-out regularization,\OursFL reduces $||h||$ and $\zeta_n^\pm$ in essence according to SVM theory, and the term above that bounds logit distance from below is hence increased. A more detailed analysis of this is given in the Appendix. 

Since projections onto the same axis are always co-linear, the common cosine similarity between them is always either 1 or -1 and thus not meaningful. We hence use Gaussian function as a similarity measurement because of its outstanding capability~\cite{yang2021parameter} as similarity kernel:
\begin{multline}
    \text{sim} (\frac{w_k ^ \tau \cdot h_{k, k'}}{||h_{k, k'}||}, \frac{w_{k'} ^ \tau \cdot h_{k, k'}}{||h_{k, k'}||}) =\\
    exp(-\frac{(w_k ^ \tau \cdot h_{k, k'} - w_{k'} ^ \tau \cdot h_{k, k'})^2}{2||h_{k, k'}||^2})
\end{multline}
\OursFL then optimizes class embeddings regarding the objective $\ell_{sp}$ and can benefit from adaptivity and momentum with a proper choice of optimizer. The max-margin spread-out regularization part of~\OursFL is illustrated in Algorithm~\ref{alg:ours_2}. The pseudocode for the whole algorithm is given in Algorithm~\ref{alg:ours_all}.

\begin{algorithm}[t]
\caption{The \OursFL Framework}\label{alg:ours_all}
\begin{algorithmic}

\State \textbf{Input:} clients $n \in [N]$, client local datasets $D_1, ..., D_N$, $|D_G|=|D_1|+...+|D_N|$, number of global epochs $T$, number of client epochs $E$, number of classes $K$, server learning rate $\eta_G$, client learning rate $\eta$, mini-batch size $B$

\hfill
\State \textbf{ServerUpdate:}
\State initialize global model $\theta_0^G = (g_0^G, W_0^G)$

\For{$t = 0, 1, ..., T-1$}
\For{client $n \in [N]$}
\State $\theta_{t+1}^n \gets$ ClientUpdate$(n, \theta_t^G)$
\EndFor
\State $g_{t+1}^G \gets \sum_{n=1}^{N} \frac{|D_n|}{|D_G|} g_{t+1}^n$
\State fit SVM using samples $\{(w_k^n, k) | k \in [K], n \in [N]\}$
\State $W_{t+1}^G \gets $ Algorithm~\ref{alg:ours_1}~\OursFL part 1
\State $W_{t+1}^G \gets $ Algorithm~\ref{alg:ours_2}~\OursFL part 2
\EndFor
\State \textbf{return} $\theta_T^G = (g_T^G, W_T^G)$

\hfill
\State \textbf{ClientUpdate$(n, \theta_t^G)$:} \Comment{Run on client $n$ with model $\theta_t^G$}
\State $\theta_{t+1}^n \gets \theta_t^G$
\For{$e = 0, 1,..., E-1$}
\For{mini-batch $\mathcal{B}$ of size $B$ in $D_n$}
\State $\theta_{t+1}^n \gets \text{client\_optimizer}(\theta_{t+1}^n, -\nabla_{\theta_{t+1}^n} \ell(\mathcal{B}), \eta)$
\EndFor
\EndFor
\State \textbf{return} $\theta_{t+1}^n$
\end{algorithmic}
\end{algorithm}

\begin{table*}[t]
    \centering
    \begin{tabular}{c c c c c c c}
        Dataset & Source & Task & \# Classes & \# Users & Type \& Dim & Mean/Std per User\\ \hline
        
        \multirow{2}{*}{FEMNIST} & \cite{lecun1998mnist} & image & \multirow{2}{*}{62} & \multirow{2}{*}{3550} & BW image & \multirow{2}{*}{226.8 / 88.9} \\
        & \cite{cohen2017emnist} & classification & & & $28 \times 28$ & \\ \hline
        
        \multirow{2}{*}{CelebA} & \multirow{2}{*}{\cite{liu2015deep}} & smile & \multirow{2}{*}{2} & \multirow{2}{*}{9343} & RGB image & \multirow{2}{*}{21.4 / 7.6}\\
        & & detection & & & $84 \times 84$ &  \\ \hline 
        
        \multirow{2}{*}{Shakespeare} & \cite{shakespeare2014complete} & next char & \multirow{2}{*}{80} & \multirow{2}{*}{1129} & string & \multirow{2}{*}{3743.2 / 6212.3}\\
        & \cite{mcmahan2017communication} & prediction & & & 80 & \\
    \end{tabular}
    \caption{Overview of used datasets.}
    \label{tab:datasets}
\end{table*}

\begin{figure*}[t]
     \centering
     \subfloat[]{\includegraphics[height=3.8cm, keepaspectratio]{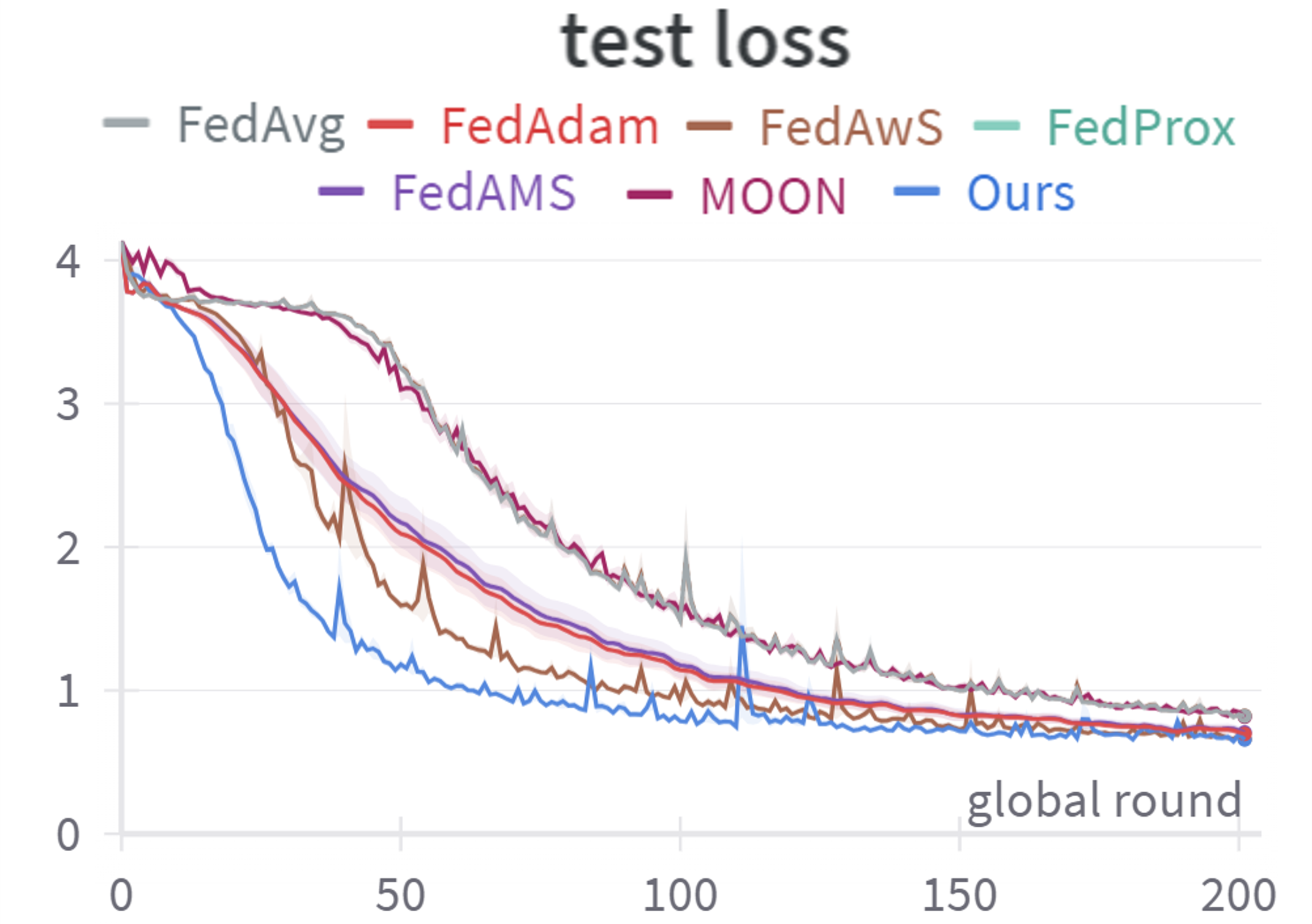}
    \label{fig:femnist_loss}}
    \hfill
    \subfloat[]{\includegraphics[height=3.8cm, keepaspectratio]{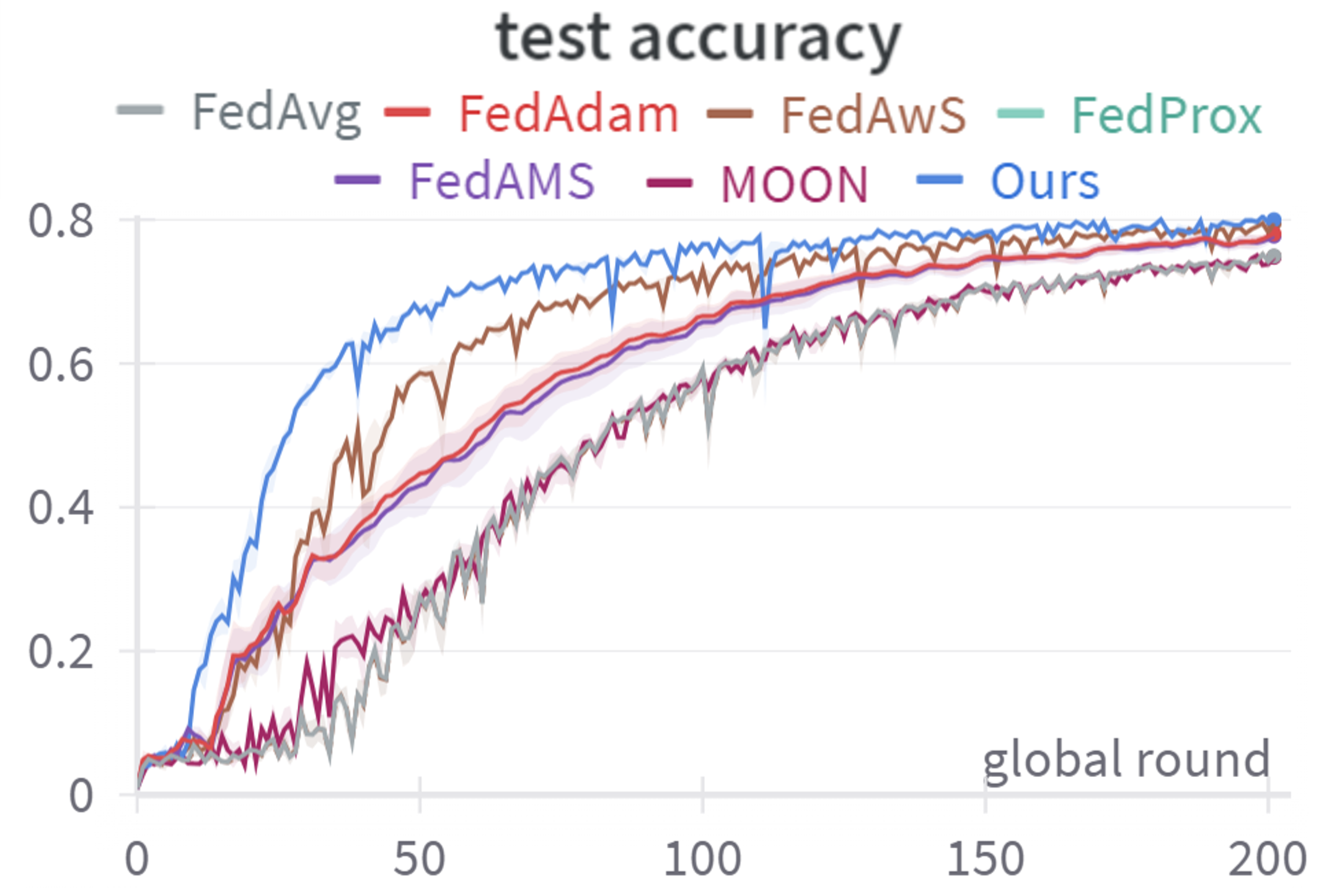}
    \label{fig:femnist_accu}}
    \hfill
    \subfloat[]{\includegraphics[height=3.8cm, keepaspectratio]{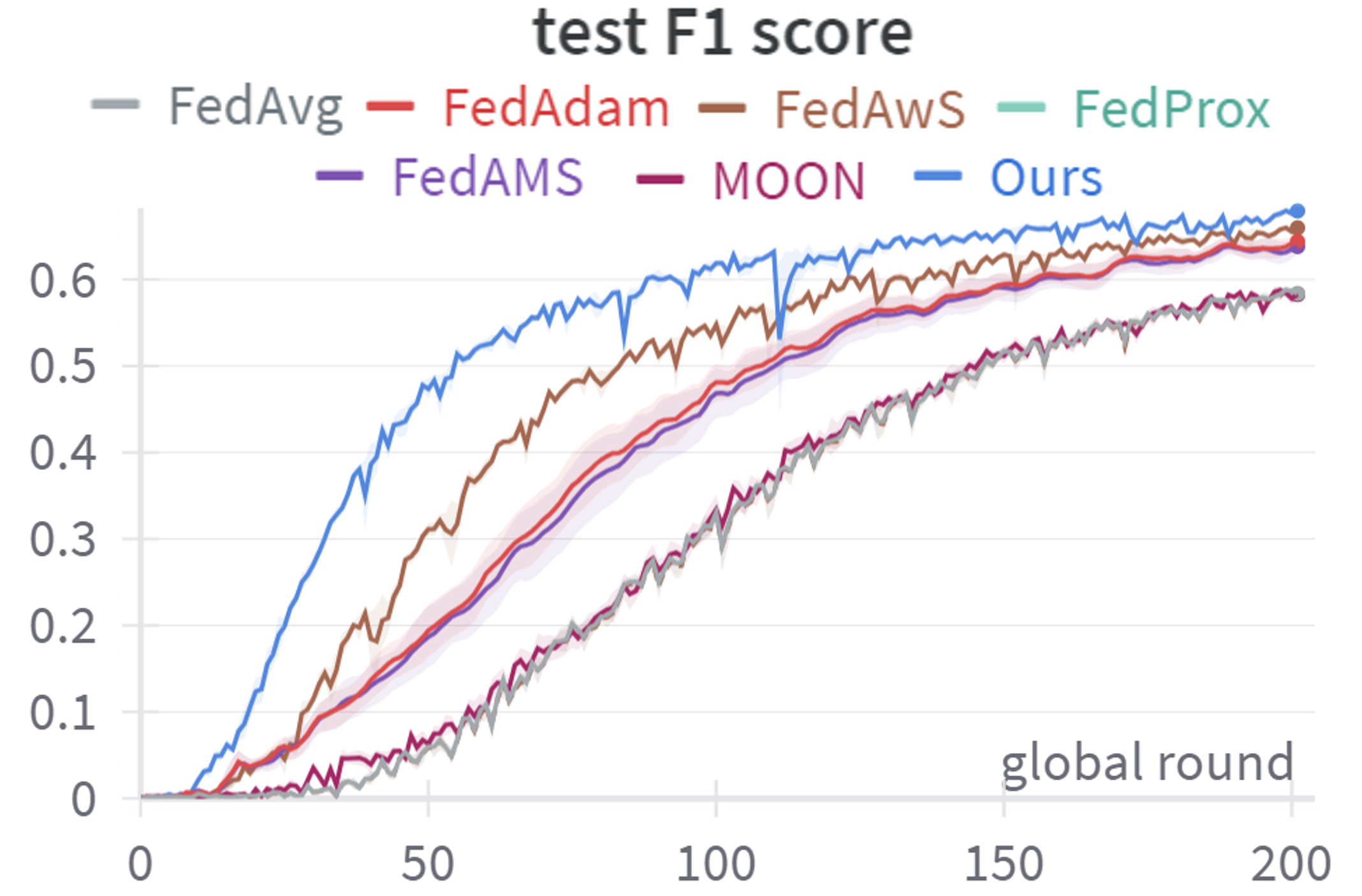}
    \label{fig:femnist_f1}}
    \hfill
    \caption{Test metrics on FEMNIST dataset.}
    \label{fig:femnist_test}
\end{figure*}


\section{Experiments and Results}
We benchmarked~\OursFL on three datasets against six FL algorithms, including FedAvg~\cite{mcmahan2017communication}, FedAdam~\cite{reddi2020adaptive}, FedAMS~\cite{wang2022communication}, FedProx~\cite{li2020federated}, MOON~\cite{li2021model}, and FedAwS~\cite{yu2020federated}. In this section, we provide task descriptions and results. For more details such as reproducibility and model structures, we redirect readers to the Appendix and our GitLab repository\footnote{~\url{https://gitlab.lrz.de/hctl/TurboSVM-FL}.}.

\subsection{Tasks}
We benchmarked~\OursFL on three different datasets covering data types of both image and nature language, namely FEMNIST~\cite{lecun1998mnist, cohen2017emnist}, CelebA~\cite{liu2015deep}, and Shakespeare~\cite{shakespeare2014complete, mcmahan2017communication} (Table~\ref{tab:datasets}). The task in FEMNIST dataset is handwritten digit and letter classification using grayscale images. The number of classes in FEMNIST is 62 (10 digits, 26 lowercase letters, and 26 uppercase letters) and the resolution of images is $28 \times 28$. The CelebA dataset contains $84 \times 84$ RGB images of faces of celebrities, and the task is binary classification between smiling and non-smiling faces. The Shakespeare dataset consists of scripts of different roles from Shakespeare's works, and the task is next-character prediction given an input string of length 80. The number of unique characters and symbols in this dataset is 80. All three datasets can be acquired on LEAF~\cite{caldas2018leaf}. For the two image classification tasks, CNN models were implemented, while for the language task LSTM model was utilized. We adopted the model structures as given in LEAF. Details about data distributions and models can be found in the Appendix.

We also adopted the data split given in LEAF. More specifically, we conducted $90\%-10\%$ train-test-split in a user-independent way, which means we had a held-out set of clients for validation rather than a fraction of validation data on each client~\cite{wang2021field}. The main reason for conducting user-independent validation is that such a test is a more valid approach for unseen data and, thus, more representative for real-world applications. Moreover, it is more challenging to fit a model in a user-independent setting compared to a user-dependent data split.

\subsection{Results}
To compare the convergence rate of different FL algorithms, we reported two groups of metrics: number of global aggregation rounds needed to reach certain validation accuracy (70\% for FEMNIST and CelebA, 50\% for Shakespeare), and the achieved F1 score, accuracy, and MCC (Matthews Correlation Coefficient) after 100 aggregation rounds. The results are given in Tables~\ref{tab:numep70} and~\ref{tab:metrics100ep}, and also visualized in Figures~\ref{fig:femnist_test}, 4 and 5 (Appendix) for each task respectively. 

\begin{table}[t]
    \centering
    \begin{tabular}{l c c c}
        Algorithm & FEMNIST & CelebA & Shakespr. \\
        \hline
        FedAvg & 144.4$\pm$4.6 & 91.6$\pm$18.2 & 51.0$\pm$5.0 \\
        FedAdam & 110.8$\pm$16.2 & $>$200 & 54.8$\pm$3.2 \\
        FedAwS & 81.4$\pm$2.2 & 84.2$\pm$24.4 & 45.0$\pm$2.3 \\
        FedProx & 145.4$\pm$3.4 & 94.4$\pm$18.4 & 157.2$\pm$5.3 \\
        FedAMS & 116.4$\pm$19.6 & $>$200 & 51.6$\pm$2.1 \\
        MOON & 145.6$\pm$3.7 & 94.2$\pm$18.8 & 52.4$\pm$3.1 \\
        \textit{\OursFLNoSpace} & \textbf{54.6$\pm$1.6} & \textbf{46.4$\pm$9.4} & \textbf{43.4$\pm$2.9} \\
    \end{tabular}
    \caption{Number of communication rounds needed to reach certain test accuracy on FEMNIST (70\%), CelebA (70\%), and Shakespeare (50\%) datasets. Smaller is better.}
    \label{tab:numep70}
\end{table}

\begin{table}[t]
    \centering
    \begin{tabular}{lcccc}
        Algorithm & [\%] & FEMN. & CelebA & Shakespr. \\
        \hline
        \multirow{3}{*}{FedAvg} & $F_1$ & 33.1$\pm$1.3 & 70.5$\pm$2.2 & 17.4$\pm$0.5 \\
        & $A$ & 59.3$\pm$1.8 & 70.6$\pm$2.1 & 52.9$\pm$0.8 \\
        & $M$ & 57.9$\pm$1.9 & 41.6$\pm$4.2 & 48.9$\pm$0.8 \\
        \hdashline
        \multirow{3}{*}{FedAdam} & $F_1$ & 48.1$\pm$4.9 & 34.2$\pm$0.3 & 18.1$\pm$0.2 \\
        & $A$ & 66.6$\pm$3.9 & 51.6$\pm$0.1 & 53.8$\pm$0.9 \\
        & $M$ & 65.4$\pm$4.1 & 1.0$\pm$2.3 & 49.8$\pm$0.9 \\
        \hdashline
        \multirow{3}{*}{FedAwS} & $F_1$ & 54.8$\pm$0.8 & 72.2$\pm$3.0 & 18.3$\pm$0.6 \\
        & $A$ & 73.1$\pm$0.6 & 72.3$\pm$2.8 & 53.3$\pm$0.7  \\
        & $M$ & 72.2$\pm$0.6 & 44.9$\pm$5.6 & 49.3$\pm$0.7 \\
        \hdashline
        \multirow{3}{*}{FedProx} & $F_1$ & 32.8$\pm$1.3 & 70.4$\pm$2.1 & 12.5$\pm$0.6 \\
        & $A$ & 59.2$\pm$1.8 & 70.6$\pm$2.1 & 46.0$\pm$1.2 \\
        & $M$ & 57.7$\pm$1.9 & 41.4$\pm$4.1 & 41.2$\pm$1.2 \\
        \hdashline
        \multirow{3}{*}{FedAMS} & $F_1$ & 46.8$\pm$6.5 & 36.3$\pm$5.0 & 18.2$\pm$0.4 \\
        & $A$ & 65.8$\pm$5.0 & 52.6$\pm$2.3 & \textbf{54.0$\pm$0.5} \\
        & $M$ & 64.6$\pm$5.2 & 4.7$\pm$10.5 & \textbf{50.0$\pm$0.6} \\
        \hdashline
        \multirow{3}{*}{MOON} & $F_1$ & 33.4$\pm$3.7 & 70.7$\pm$2.3 & 17.6$\pm$0.6 \\
        & $A$ & 58.2$\pm$3.0 & 70.9$\pm$2.2 & 52.8$\pm$1.0 \\
        & $M$ & 56.8$\pm$3.1 & 41.8$\pm$4.4 & 48.8$\pm$1.0 \\
        \hdashline
        \multirow{3}{*}{\makecell{\textit{TurboSVM-} \\ \textit{FL}}} & $F_1$ & \textbf{61.9$\pm$0.7} & \textbf{77.2$\pm$1.3} & \textbf{19.2$\pm$0.3} \\
        & $A$ & \textbf{76.6$\pm$1.0} & \textbf{77.3$\pm$1.2} & 53.7$\pm$0.2 \\
        & $M$ & \textbf{75.8$\pm$1.0} & \textbf{55.0$\pm$2.4} & 49.7$\pm$0.2 \\
    \end{tabular}
    \caption{Achieved test F1 score, accuracy (A), and MCC score (M) after 100 aggregation rounds. Greater is better.}
    \label{tab:metrics100ep}
\end{table}

Our results clearly indicate that on FEMNNIST and CelebA datasets~\OursFL yields a significantly faster convergence in contrast to other FL methods, while on the Shakespeare dataset~\OursFL slightly outperforms others. Compared to the baseline FedAvg,~\OursFL successfully reduces the number of global rounds by 62.2\%, 49.3\%, and 14.9\% to reach the same test accuracy as given in Table~\ref{tab:numep70}. When all FL methods are run for the same rounds,~\OursFL yields in much better test metrics on both image classification tasks in comparison to other methods, while its performance is a fair match to the adaptive algorithms on the next-character prediction task. Moreover, we show in Figure 4 that while adaptive FL methods like FedAdam and FedAMS are not stable,~\OursFL can still robustly benefit from adaptivity on the server side.

\subsection{Impact of Embedding Size on Selectivity}
We further explored the impact of embedding size on the number of client models that form support vectors. To approach this, we ran experiments on the FEMNIST dataset with varying embedding size and number of participating clients $C$. We recorded the number of support vectors for class 1 at 200th round in Table~\ref{tab:ablation_embed_size}. 
\begin{table}[b]
    \centering
    \begin{tabular}{c c|c c c c c}
        \multicolumn{2}{c}{ } & \multicolumn{5}{c}{Embedding Dimension} \\
          $C$ & & 4 & 16 & 64 & 256 & 1024 \\
          \hline
         \multirow{2}{*}{8} & \#SV & 8 & 8 & 8 & 8 & 8 \\ 
          & $F_1$[\%] & 6.1 & 14.2 & 58.9 & 63.9 & 67.4 \\ 
          \hline
         \multirow{2}{*}{64} & \#SV & 62 & 63 & 58 & 52 & 37 \\
          & $F_1$[\%] & 16.3 & 43.2 & 60.5 & 65.9 & 68.3 \\ 
          \hline
         \multirow{2}{*}{512} & \#SV & 197 & 150 & 147 & 100 & 77 \\ 
          & $F_1$[\%] & 16.7 & 43.2 & 59.7 & 64.5 & 67.0 \\ 
    \end{tabular}
    \caption{Influence of embedding size on support vectors.}
    \label{tab:ablation_embed_size}
\end{table}
It is clear that a higher embedding size is associated with better performance and fewer support vectors when there exist enough participating clients, while a scarcity of clients can lead to full use of class embeddings as support vectors. Furthermore, larger embedding size also leads to higher complexity and burden, which needs to be balanced off depending on the specific tasks.

\section{Discussion}
As~\OursFL focuses on class representations on the server side while other parts of the global model are still aggregated with average and client training is done with vanilla SGD, the use of~\OursFL in combination with other FL algorithms is promising. For instance, on the client side FedProx can be applied to counteract data heterogeneity, while on the server side, class embeddings are aggregated with~\OursFLNoSpace. Another example is the use of adaptive FL methods like FedAdam for encoder aggregation while logit layers are aggregated with~\OursFLNoSpace. Moreover the idea of model-as-sample can be further explored, for example, for anomaly client detection and client clustering.

~\OursFL is particularly suitable for cross-device FL where edge devices are often constrained by computation and storage resources, but its improvement is also not excluded from cross-silo case. A typical application scenario of~\OursFL is federated transfer learning, where a pre-trained model like VGG16 and Resnet50 is adopted, and all of the layers except the last few ones are frozen. In this case, each client only needs to train and share the last few layers, which makes~\OursFL extremely efficient. The capability of~\OursFL is also not constrained to single-output tasks. For multi-output tasks such as multi-label classification and multi-task learning,~\OursFL can also be applied. To approach this, separate classification heads for different tasks should be implemented where the backbone encoder shares its weights among tasks, and then TurboSVM-FL should be applied to each head in parallel.

One improvement direction of~\OursFL is to relax the implicit assumption about linear separability of class embeddings with kernelization during SVM fitting and class inference. A piloting ablation study is included in the Appendix in this regard. Furthermore, while posing no additional computation cost on the client side,~\OursFL requires the server to be powerful such that it can fit SVMs efficiently, especially when the SVMs are in one-vs-one (OVO) pattern. We chose OVO instead of OVR (one-vs-rest) mainly for two reasons: 1. in general, OVO performs better than OVR; 2. for TurboSVM-FL, OVO never suffers from class imbalance while OVR always does, since the numbers of samples for each class are always the same. Although OVO imposes more computation on the server side, we think that to approach FL, a powerful server is a must-have, and OVO is no burden for such a server. In case the number of classes is large, the computation burden can be further resolved by sampling a proportion of classes on which SVMs are fitted. 

\section{Conclusion}
In this work, we proposed a novel federated aggregation strategy called~\OursFLNoSpace, which extensively exploits SVM to conduct selective aggregation and max-margin spread-out regularization for class embeddings and can vastly reduce communication rounds. We tested our approach on three publicly available datasets, and our results show that~\OursFL outperforms existing FL methods largely on convergence rate regarding various metrics.

\section*{Acknowledgements}
We acknowledge the funding by the Deutsche Forschungsgemeinschaft (DFG, German Research Foundation) – Project number KA 4539/5-1.


\bibliography{aaai24}

\appendix
\onecolumn
\section{Appendix}\label{chap:appendix}

\hspace{1cm}
\subsection{\centering Implementation}
Our implementation of~\OursFL and instructions for reproducing experiment results can be found on our GitLab repository (\url{https://gitlab.lrz.de/hctl/TurboSVM-FL}).

\hspace{1cm}
\subsection{\centering Analysis}
It is hard to give an analysis to the convergence rate of~\OursFL directly. Instead, we provide informal descriptive analysis to how~\OursFL is potential to reduce misclassification error. In~\cite{yu2020federated} it is proven that misclassification error is related to the separation of class embeddings and that the cosine contrastive loss with spread-out regularization surrogates misclassification error, which inspire our work. For detailed proof in this direction, we redirect readers to~\cite{yu2020federated}. 

To simplify the analysis, we narrow down to binary classification without loss of generality and ignore all bias terms. In a federated classification task with $N$ clients, denote the class embeddings of positive and negative classes as $x_1^+, ..., x_N^+$ and $x_1^-, ..., x_N^-$ respectively. Let $I_n^+ \in \{0, 1\}, I_n^- \in \{0, 1\}$ be the indicator of whether a class embedding is used as support vector and $|D_n|$ denote the sample size of client $n$. Then, the aggregated class embeddings can be given as $\frac{\sum_{n=1}^{N}I_n^+ |D_n| x_n^+}{\sum_{n=1}^{N}I_n^+ |D_n|}$ and $\frac{\sum_{n=1}^{N}I_n^- |D_n| x_n^-}{\sum_{n=1}^{N}I_n^- |D_n|}$. Let $h$ be the decision boundary of SVM fitted on $\{(x_n^+, +) | n \in [N] \} \cup \{(x_n^-, -) | n \in [N] \}$ and $\zeta_1^+, ..., \zeta_N^+, \zeta_1^-, ..., \zeta_N^-$ be the corresponding SVM relaxation terms, i.e.:
\begin{gather*}
    h, \zeta_1^+, ..., \zeta_N^+, \zeta_1^-, ..., \zeta_N^- = \argmin_{h, \zeta_1^\pm, ..., \zeta_N^\pm} \frac{1}{2} ||h||^2 + \lambda \sum_{n=1}^{N} (\zeta_n^+ + \zeta_n^-), \\
    \text{s.t. } h^\tau \cdot x_n^+ \geq 1 - \zeta_n^+, h^\tau \cdot x_n^- \leq -(1 - \zeta_n^-) \text{ and } \zeta_n^+, \zeta_n^- \geq 0
    \label{eq:svm2}
\end{gather*}
where the first objective term $\frac{1}{2} ||h||^2$ corresponds to margin maximization and the second term allows each sample to be away from its correct margin up to distance $\zeta_n$. Given a new sample with extracted feature $x^*$, without loss of generality, assume its true label is positive. Further, we assume $x^*$ is a ``good" sample and can be not only correctly but also ``well" classified, which means $h^\tau \cdot x^* \geq 1 - \zeta^*$ with $\zeta^* \leq 1$. Let $\zeta_n^+, \zeta_n^- \leq 1$ for all $n \in [N]$. We then take a look into the difference between the positive and negative logits when class embeddings are projected onto SVM separating hyperplane, i.e.:
\begin{align*} 
 & \text{proj(logit$^+(x^*)$, $h$)} - \text{proj(logit$^-(x^*)$, $h$)} \\
=& \frac{(\sum_{n=1}^{N}I_n^+ |D_n| x_n^+)^\tau \cdot h}{\sum_{n=1}^{N}I_n^+ |D_n|||h||} \frac{h^\tau}{||h||} \cdot x^* - \frac{(\sum_{n=1}^{N}I_n^- |D_n| x_n^-)^\tau \cdot h}{\sum_{n=1}^{N}I_n^- |D_n|||h||} \frac{h^\tau}{||h||} \cdot x^* \\
=& \frac{\lambda^- \sum_{n=1}^{N}I_n^+ |D_n| (x_n^{+\tau} \cdot h) - \lambda^+ \sum_{n=1}^{N}I_n^- |D_n| (x_n^{-\tau} \cdot h)}{\lambda^+\lambda^-||h||^2} h^\tau \cdot x^* \hspace{1.0cm} (\lambda^+ = \sum_{n=1}^{N}I_n^+ |D_n|, \lambda^- = \sum_{n=1}^{N}I_n^- |D_n|) \\
\geq& \frac{2\lambda^+\lambda^- - \sum_{n=1}^{N}(\lambda^- I_n^+|D_n|\zeta_n^+ - \lambda^+ I_n^-|D_n|\zeta_n^-)}{\lambda^+ \lambda^- ||h||^2}(h^\tau \cdot x^*) \hspace{5.3cm} \text{(SVM property)} \\
\geq& \frac{[2\lambda^+\lambda^- - \sum_{n=1}^{N}(\lambda^- I_n^+|D_n|\zeta_n^+ - \lambda^+ I_n^-|D_n|\zeta_n^-)](1 - \zeta^*)}{\lambda^+ \lambda^- ||h||^2} \hspace{4.4cm} \text{($x^*$ a ``good" sample)} \\
\geq& \frac{[2N - \sum_{n=1}^{N}(\zeta_n^+ + \zeta_n^-)](1 - \zeta^*)}{N ||h||^2} \hspace{3.2cm} \text{(assume $I_n^\pm = 1$ for all $n \in [N]$ and $|D_1| = ... = |D_N|$)}
\end{align*}
When we optimize the aggregated class embeddings $\frac{\sum_{n=1}^{N}I_n^+ |D_n| x_n^+}{\sum_{n=1}^{N}I_n^+ |D_n|}, \frac{\sum_{n=1}^{N}I_n^- |D_n| x_n^-}{\sum_{n=1}^{N}I_n^- |D_n|}$ with max-margin spread-out regularization in~\OursFLNoSpace, we reduce $||h||$ and $\zeta_n^\pm$ in essence according to SVM theory, and the term above that bounds logit distance from below is hence increased. As discovered in~\cite{yu2020federated}, a larger distance between class embeddings or logits has the capability to reduce the probability of misclassification. Therefore, the max-margin spread-out regularization on class embeddings is potential to reduce misclassification error when class embeddings are projected onto SVM decision boundary $w$. The effect also propagates to the case when class embeddings are not projected onto $h$ given $h^\tau \cdot x^* \geq 0$. 

\hspace{1cm}
\subsection{\centering Experiment Environment}
We implemented~\OursFL and other FL methods with PyTorch and ran experiments on multiple computers. All computers are equipped with exactly the same hardware (32 GB RAM, Intel i7-13700K, NVIDIA RTX 4080 16 GB), operating system (WSL2 Ubuntu 22.04.2 LTS), and software (Python 3.10.12, PyTorch 2.0.1 for CUDA 11.7). 

\hspace{1cm}
\subsection{\centering Randomness}
All our experiments were replicated five times with different random seeds from $\{0, 1, 2, 3, 4\}$ each time. The random seed applies to $numpy.random.seed$, $torch.manual\_seed$, and $random.seed$ to guarantee reproducibility. We reported the mean and standard deviation (std) over five seeds for each metric. 

\hspace{1cm}
\subsection{\centering Data Distribution}
\begin{figure*}[!htb]
     \centering
     \subfloat[FEMNIST dataset]{\includegraphics[height=4.2cm, keepaspectratio]{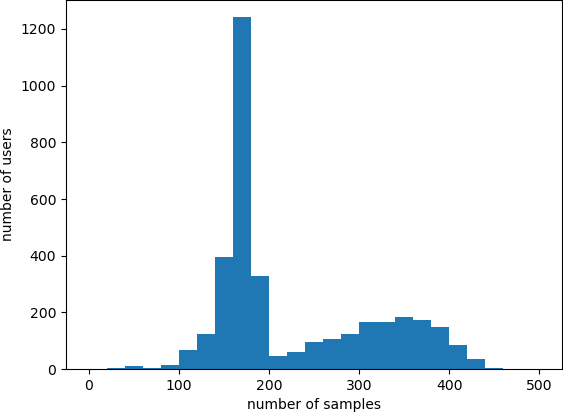}
    \label{fig:femnist_histo}}
    \subfloat[Celeba dataset]{\includegraphics[height=4.2cm, keepaspectratio]{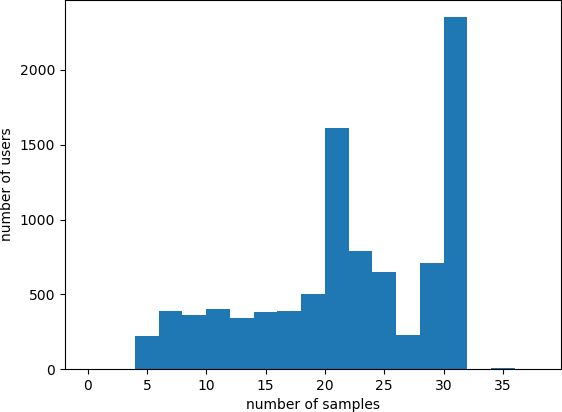}
    \label{fig:celeba_histo}}
    \subfloat[Shakespeare dataset]{\includegraphics[height=4.2cm, keepaspectratio]{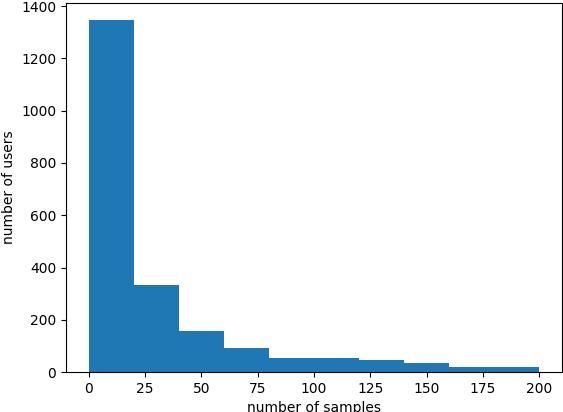}
    \label{fig:shakespeare_histo}}
    \hfill
    \caption{Histogram of number of samples per user in the datasets from LEAF~\cite{caldas2018leaf}.}
    \label{fig:histo1}
\end{figure*}

The histograms of number of samples per client of each dataset are given in Figure~\ref{fig:histo1}.

\hspace{1cm}
\subsection{\centering Model Architectures}
The FL benchmark framework LEAF~\cite{caldas2018leaf} provides standard model in Tensorflow for each task. We translated all these models into PyTorch and kept their architectures as given in LEAF. Detailed model architectures are described in Tables~\ref{tab:femnist_model},~\ref{tab:celeba_model},~\ref{tab:shakespeare_model} in the Appendix. For all tasks, the activation function for logit layer is softmax, while the classification objective is cross entropy, regardless of whether the classification task is binary or multi-class.
\begin{table}[!htb]
    \centering
    \begin{tabular}{l l}
        Layer & Architecture \\ \hline
        Input & shape $1 \times 28 \times 28$ \\
        Conv2d & kernel size 5, in/out channel 1/32, same padding \\ 
        ReLU & - \\
        MaxPooling & kernel size 2, stride 2 \\
        Conv2d & kernel size 5, in/out channel 32/64, same padding \\ 
        ReLU & - \\
        MaxPooling & kernel size 2, stride 2 \\
        Flatten & - \\
        Linear & in/out dimension 3136/2048 \\
        ReLU & - \\
        Linear & in/out dimension 2048/62 \\
    \end{tabular}
    \caption{Model structure (CNN) for FEMNIST dataset, following~\url{https://github.com/TalwalkarLab/leaf/blob/master/models/femnist/cnn.py}.}
    \label{tab:femnist_model}
\end{table}

\begin{table}[!htb]
    \centering
    \begin{tabular}{l l}
        Layer & Architecture \\ \hline
        Input & shape $3 \times 84 \times 84$ \\
        Conv2d & kernel size 3, in/out channel 3/32, same padding \\ 
        BatchNorm2d & - \\
        MaxPolling & kernel size 2, stride 2 \\
        ReLU & - \\
        
        Conv2d & kernel size 3, in/out channel 32/32, same padding \\ 
        BatchNorm2d & - \\
        MaxPolling & kernel size 2, stride 2 \\
        ReLU & - \\
        
        Conv2d & kernel size 3, in/out channel 32/32, same padding \\ 
        BatchNorm2d & - \\
        MaxPolling & kernel size 2, stride 2 \\
        ReLU & - \\
        
        Conv2d & kernel size 3, in/out channel 32/32, same padding \\ 
        BatchNorm2d & - \\
        MaxPolling & kernel size 2, stride 2 \\
        ReLU & - \\
        
        Flatten & - \\
        Linear & in/out dimension 800/2 \\ 
    \end{tabular}
    \caption{Model structure (CNN) for CelebA dataset, following~\url{https://github.com/TalwalkarLab/leaf/blob/master/models/celeba/cnn.py}.}
    \label{tab:celeba_model}
\end{table}

\begin{table}[!htb]
    \centering
    \begin{tabular}{l l}
        Layer & Architecture \\ \hline
        Embedding & number of embeddings 80, dimension 8 \\
        LSTM & 3n/hidden dimension 8/256, hidden layers 2 \\ 
        Linear & in/out dimension 256/80 \\
    \end{tabular}
    \caption{Model structure (LSTM) for Shakespeare dataset, following~\url{https://github.com/TalwalkarLab/leaf/blob/master/models/shakespeare/stacked_lstm.py}.}
    \label{tab:shakespeare_model}
\end{table}

\hspace{1cm}
\subsection{\centering Hyperparameters}
In the process of hyperparameter tuning and experimentation, we set the client epoch $E$ to 1 and the number of participating clients $C$ to 8 by default to simulate ``lazy" clients. We also ran experiments with $E=2, 4$ and $C=16, 32$. The results of these experiments can be found in the Appendix. 

One key hyperparameter for~\OursFL is the regularization coefficient $\lambda$ for SVM fitting. We tested three different strategies for this coefficient, namely linearly increasing, linearly decreasing, and constant. Among these three strategies, linearly increasing yields the best model performance. Our explanation to this phenomenon is that as training procedure progresses, more client models approach their optima and hence become more informative. A decreasing regularization factor tends to result in an increasing number of support vectors, which matches the increase of client model informativity. We then implemented this strategy as default. Specifically, the regularization coefficient is set to $1.0$ in the beginning and is successively reduced in each global aggregation round. 

Another SVM-related factor is whether the SVMs are fitted in one-vs-one (OVO) pattern or one-vs-rest (OVR) pattern. By default, our method trains SVM in OVO pattern, which means that one binary SVM is trained for each class pair, and in total, $\frac{K(K-1)}{2}$ SVMs are trained for a multi-classification task with $K$ classes. We choose OVO instead of OVR mainly for two reasons. Firstly, OVO performs better than OVR in general. Secondly, for~\OursFL, OVO never suffers from class imbalance while OVR always does, since the numbers of samples for each class in the higher level SVM problem are always the same. Although OVO imposes more computation on the server side, we think that to approach FL, a powerful server is a must-have, and OVO is no burden for such a server. In case the number of classes is large, the computation burden can be further resolved by sampling a proportion of classes on which SVMs are fitted.

When implementing all seven FL algorithms, we followed the recommendations provided in~\cite{reddi2020adaptive, wang2021field} on the choice of optimizers, namely: the optimizer on the client side is SGD, while Adam is applied on the server side for all methods that require a server-level optimizer during central aggregation, including FedAdam, FedAMS, FedAwS, and~\OursFLNoSpace. For each task, we first ran a grid search for optimal client learning rates in $\{1e^{-5}, 1e^{-4}, 1e^{-3}, 1e^{-2}, 1e^{-1}, 1e^{0}\}$ for FedAvg. Then, we fixed the client learning rate to its optima and conducted a grid search for optimal global learning rates in the same range. As FedAMS is an improved version of FedAdam, we applied the same global learning rate for them. The details of learning rates are listed in Table~\ref{tab:lr} in the Appendix. Moreover, we decided the sizes of mini-batch based on the sample histograms (Figure \ref{fig:histo1}). The final batch sizes are 64 for FEMNIST, 8 for CelebA, and 64 for Shakespeare. The coefficients for additional penalty terms were set to $0.01$ and $1$ respectively for FedProx and MOON.

\begin{table}[!htb]
    \centering
    \begin{tabular}{l c c c }
        Algorithm & FEMNIST & CelebA & Shakespeare  \\
        \hline
        FedAvg & $1e-1$ & $1e-3$ & $1e0$ \\
        FedAdam & $1e-3$ & $1e-3$ & $1e-2$ \\
        FedAMS & $1e-3$ & $1e-3$ & $1e-2$ \\
        FedProx & $1e-1$ & $1e-3$ & $1e0$ \\
        MOON & $1e-1$ & $1e-3$ & $1e0$ \\
        FedAwS & $1e-2$ & $1e-2$ & $1e0$ \\
        \OursFLNoSpace & $1e-2$ & $1e-2$ & $1e0$ \\
    \end{tabular}
    \caption{Learning rates used in the experiments. All methods share the same client learning rates as FedAvg. For FedAvg, FedProx, and MOON, client learning rate is listed in the table, while for FedAdam, FedAwS, FedAMS, and~\OursFLNoSpace, server learning rate is given.}
    \label{tab:lr}
\end{table}

\hspace{1cm}
\subsection{\centering Additional Experiment Results}
\begin{figure*}[!htb]
     \centering
     \subfloat[]{\includegraphics[height=4.1cm, keepaspectratio]{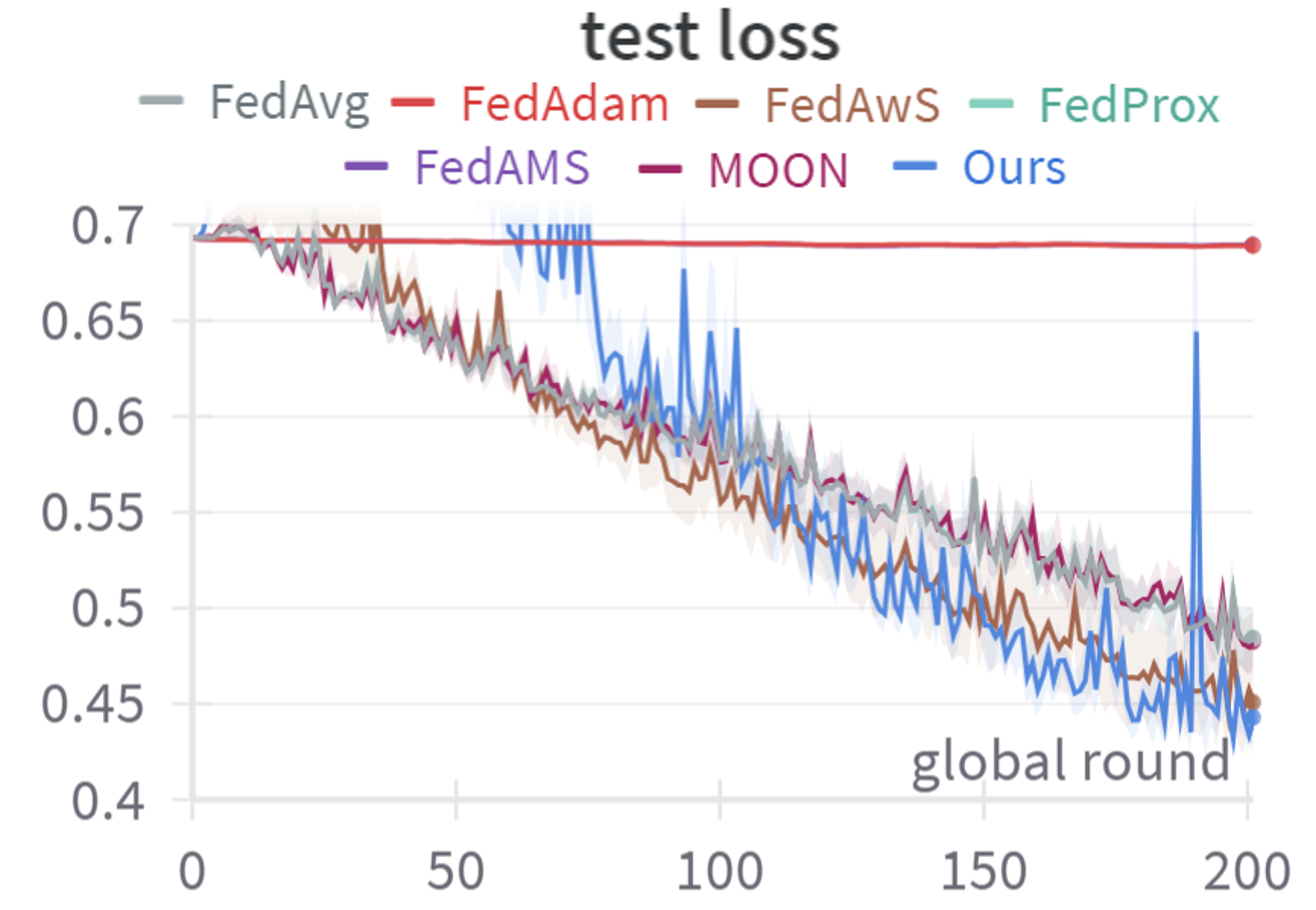}
    \label{fig:celeba_loss}}
    \subfloat[]{\includegraphics[height=4.1cm, keepaspectratio]{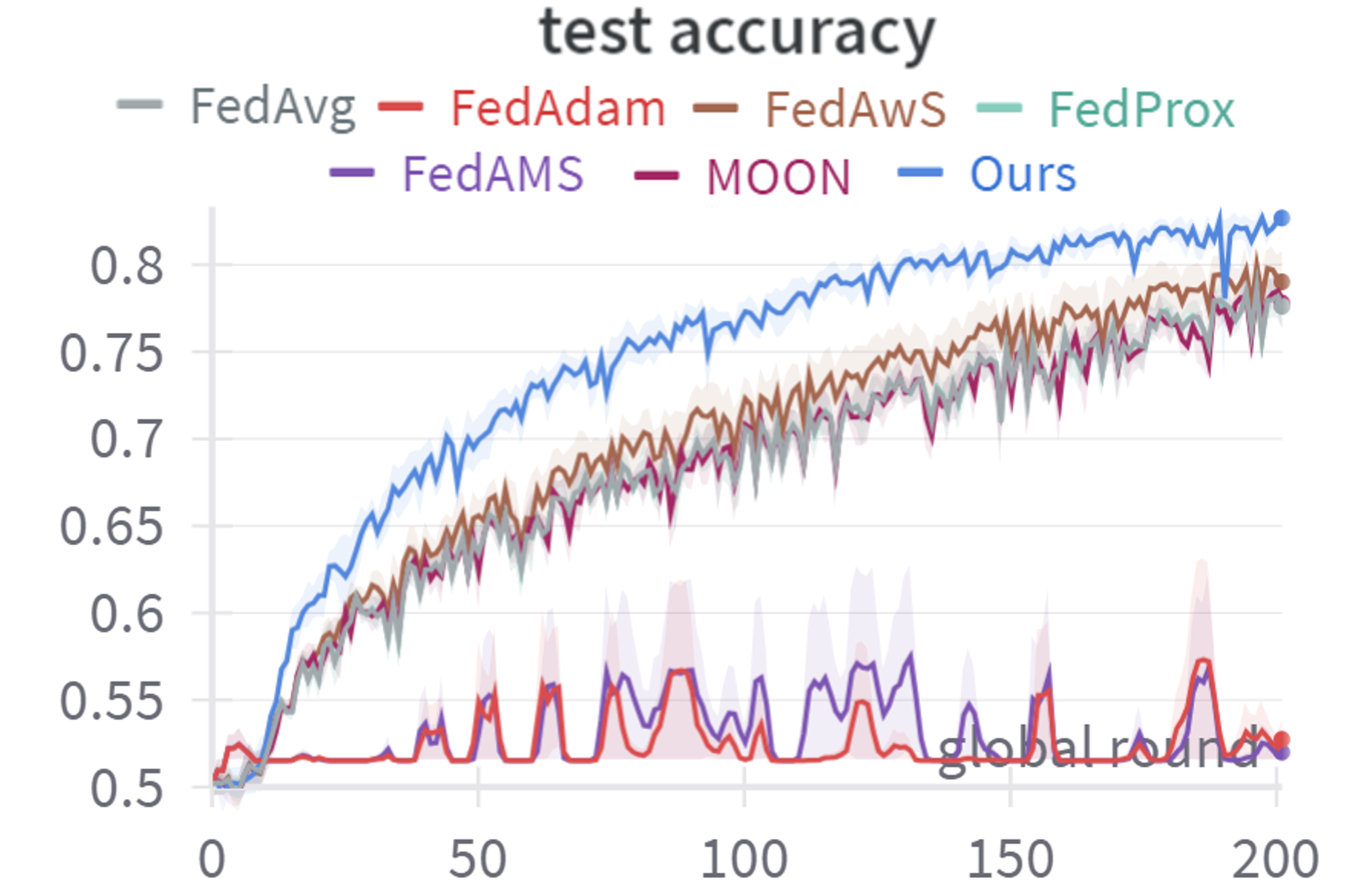}
    \label{fig:celeba_accu}}
    \subfloat[]{\includegraphics[height=4.1cm, keepaspectratio]{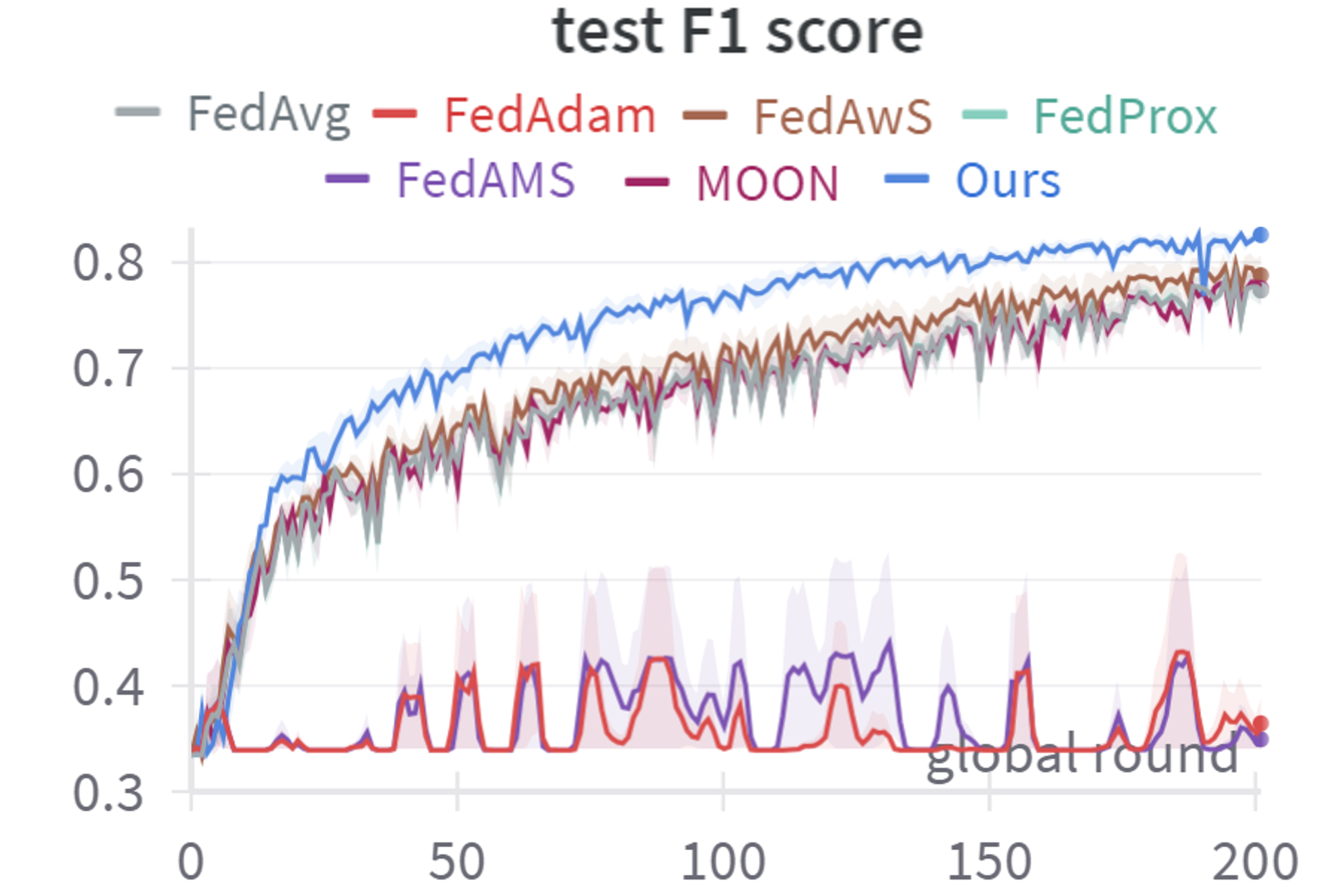}
    \label{fig:celeba_f1}}
    \hfill
    \caption{Test metrics on CelebA dataset.}
    \label{fig:celeba_test}
\end{figure*}

\begin{figure*}[!htb]
     \centering
     \subfloat[]{\includegraphics[height=4.1cm, keepaspectratio]{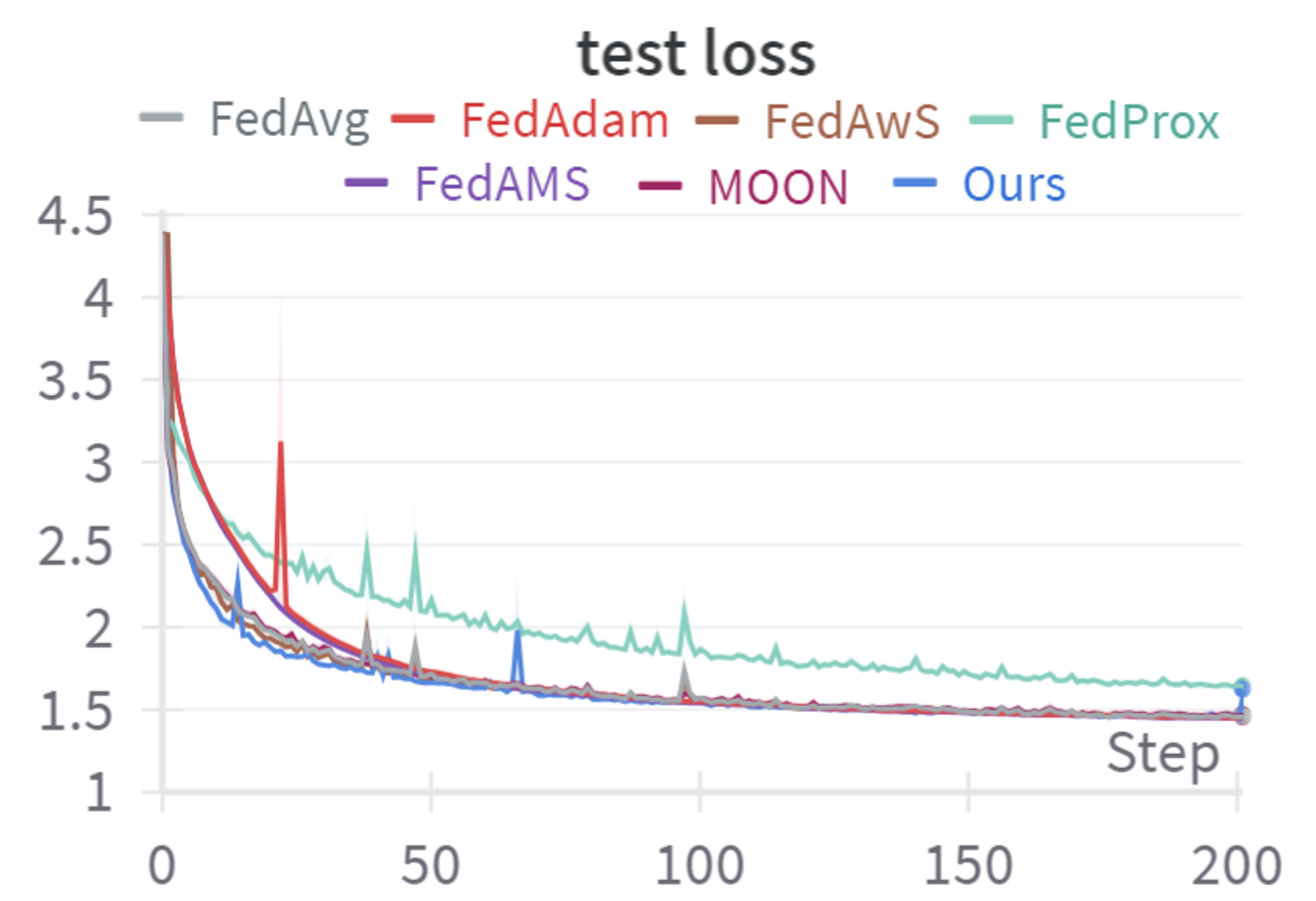}
    \label{fig:shakespeare_loss}}
    \subfloat[]{\includegraphics[height=3.9cm, keepaspectratio]{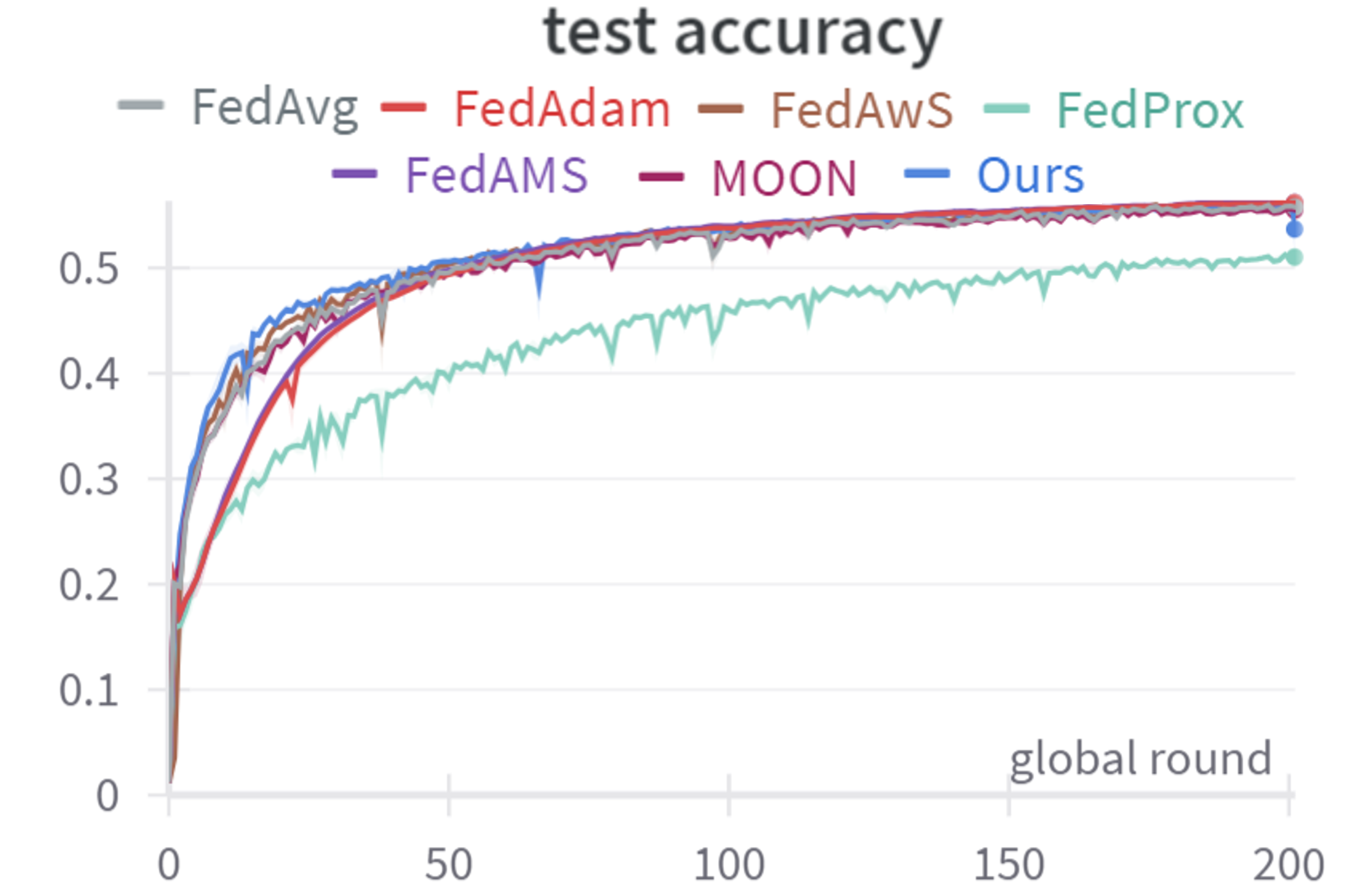}
    \label{fig:shakespeare_accu}}
    \subfloat[]{\includegraphics[height=4.1cm, keepaspectratio]{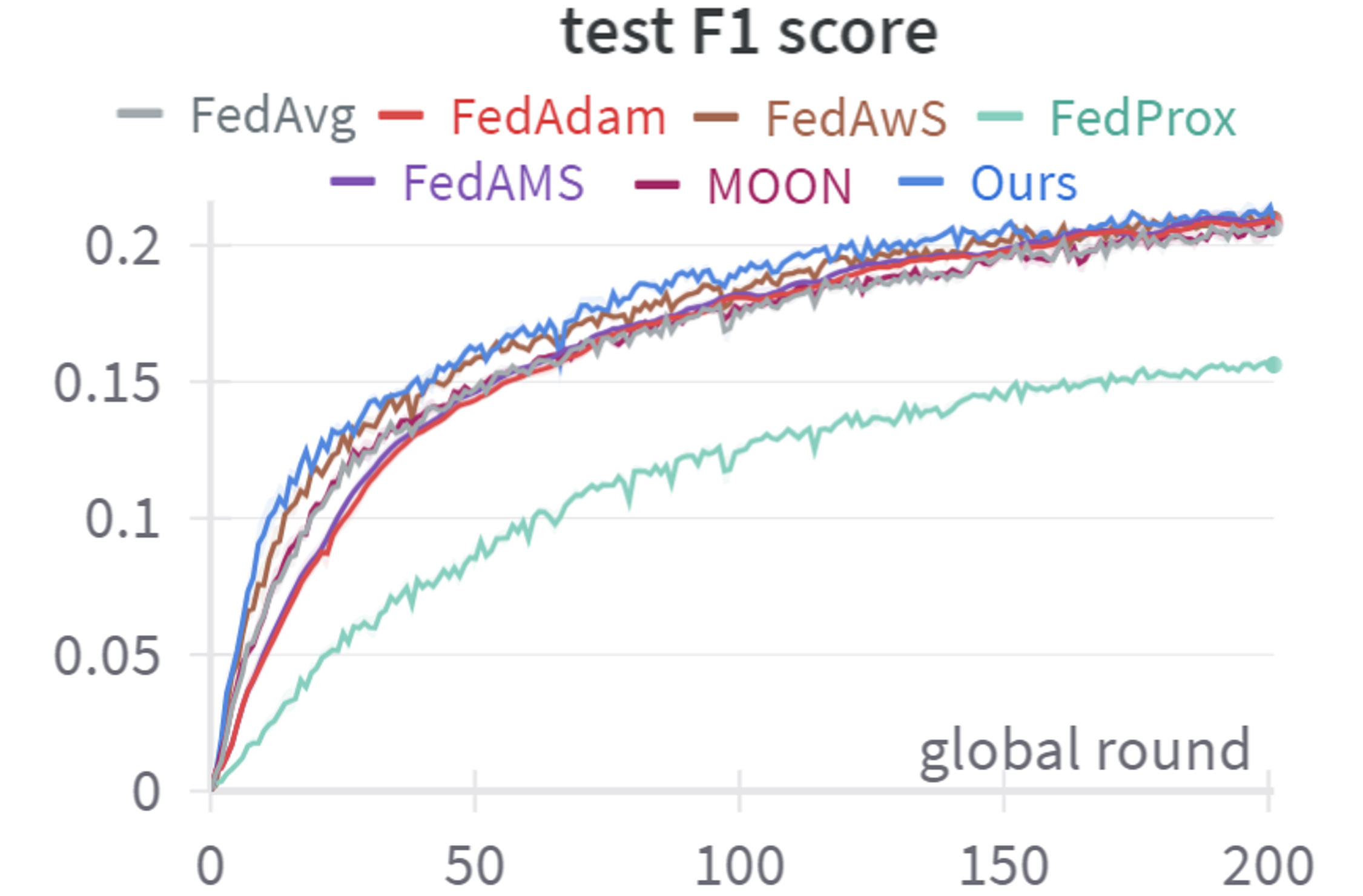}
    \label{fig:shakespeare_f1}}
    \hfill
    \caption{Test metrics on Shakespeare dataset.}
    \label{fig:shakespeare_test}
\end{figure*}

The results for the experiments using the CelebA dataset and the Shakespeare dataset with default settings ($C=8, E=1$) are plotted in Figures~\ref{fig:celeba_test} and~\ref{fig:shakespeare_test}. The results corresponding to varying number of participating clients ($C$) are given in Tables~\ref{tab:numep70_2} and~\ref{tab:metrics100ep_2}. The results regarding varying number of client local training epochs ($E$) are given in Table~\ref{tab:metrics100ep_3}.

\begin{table*}[!htb]
    \centering
    \begin{tabular}{l | c c c c c c c c c }
        Algorithm & \multicolumn{3}{c}{FEMNIST} & \multicolumn{3}{c}{CelebA} & \multicolumn{3}{c}{Shakespeare}  \\
        \#clients (C)& 8 & 16 & 32 & 8 & 16 & 32 & 8 & 16 & 32  \\
        \hline
        FedAvg & 144 & 140 & 136 & 92 & 97 & 90 & 51 & 38 & 32 \\
        FedAdam & 111 & 95 & 83 & $>$200 & $>$200 & $>$200 & 55 & 41 & 38  \\
        FedAwS & 81 & 75 & 72 & 84 & 77 & 74 & 45 & \textbf{34} & \textbf{28}  \\
        FedProx & 145 & 140 & 136 & 94 & 97 & 90 & 157 & 137 & 129  \\
        FedAMS & 116 & 97 & 83 & $>$200 & $>$200 & $>$200 & 52 & 43 & 36  \\
        MOON & 146 & 140 & 135 & 94 & 97 & 91 & 52 & 42 & 33   \\
        \textit{\OursFLNoSpace} & \textbf{55} & \textbf{51} & \textbf{54} & \textbf{46} & \textbf{44} & \textbf{40} & \textbf{43} & \textbf{34} & \textbf{28}  \\
    \end{tabular}
    \caption{Number of communication rounds needed to reach certain test accuracy on FEMNIST (70\%), CelebA (70\%), and Shakespeare (50\%) datasets respectively.}
    \label{tab:numep70_2}
\end{table*}

\begin{table*}[!htb]
    \centering
    \begin{tabular}{l c | c c c c c c c c c }
        \multicolumn{2}{l}{Algorithm} & \multicolumn{3}{c}{FEMNIST} & \multicolumn{3}{c}{CelebA} & \multicolumn{3}{c}{Shakespeare}  \\
        & [\%] / C & 8 & 16 & 32 & 8 & 16 & 32 & 8 & 16 & 32 \\
        \hline
        \multirow{3}{*}{FedAvg} & $F_1$ & 33.1 & 34.8 & 36.4 & 70.5 & 68.9 & 70.5 & 17.4 & 18.9 & 20.0   \\
        & $Acc$ & 59.3 & 60.7 & 61.7 & 70.6 & 69.4 & 70.6 & 52.9 & 54.8 & 55.9   \\
        & $MCC$ & 57.9 & 59.2 & 60.3 & 41.6 & 40.2 & 41.5 & 48.9 & 51.0 & 52.1 \\
        \hdashline
        \multirow{3}{*}{FedAdam} & $F_1$ & 48.1 & 54.8 & 58.2 & 34.2 & 37.1 & 40.0 & 18.1 & 19.6 & 20.2   \\
        & $Acc$ & 66.6 & 71.2 & 74.4 & 51.6 & 52.7 & 54.5 & 53.8 & \textbf{55.4} & \textbf{56.2}   \\
        & $MCC$ & 65.4 & 69.9 & 73.1 & 1.0 & 5.8 & 7.0 & 49.8 & \textbf{51.6} & 52.5\\
        \hdashline
        \multirow{3}{*}{FedAwS} & $F_1$ & 54.8 & 57.3 & 58.3 & 72.2 & 70.6 & 72.4 & 18.3 & 19.5 & 20.8 \\
        & $Acc$ & 73.1 & 74.6 & 74.9 & 72.3 & 71.2 & 72.5 & 53.3 & 55.1 & 56.1   \\
        & $MCC$ & 72.2 & 73.7 & 74.0 & 44.9 & 43.6 & 45.3 & 49.3 & 51.3 & 52.3  \\
        \hdashline
        \multirow{3}{*}{FedProx} & $F_1$ & 32.8 & 34.5 & 36.2 & 70.4 & 68.9 & 70.5 & 12.5 & 13.2 & 13.8   \\
        & $Acc$ & 59.2 & 60.5 & 61.6 & 70.6 & 69.4 & 70.6 & 46.0 & 47.7 & 48.2   \\
        & $MCC$ & 57.7 & 59.0 & 60.2 & 41.4 & 40.1 & 41.6 & 41.2 & 43.1 & 43.7 \\
        \hdashline
        \multirow{3}{*}{FedAMS} & $F_1$ & 46.8 & 54.4 & 59.6 & 36.3 & 37.0 & 34.9 & 18.2 & 19.5 & 20.5  \\
        & $Acc$ & 65.8 & 70.9 & 74.3 & 52.6 & 52.6 & 51.9 & \textbf{54.0} & \textbf{55.4} & \textbf{56.2}  \\
        & $MCC$ & 64.6 & 69.9 & 73.4 & 4.7 & 6.2 & 2.3 & \textbf{50.0} & \textbf{51.6} & \textbf{52.7} \\
        \hdashline
        \multirow{3}{*}{MOON} & $F_1$ & 33.4 & 36.7 & 37.6 & 70.7 & 69.8 & 70.8 & 17.6 & 18.8 & 20.0   \\
        & $Acc$ & 58.2 & 61.1 & 61.96 & 70.9 & 70.2 & 70.8 & 52.8 & 54.7 & 55.7   \\
        & $MCC$ & 56.8 & 59.7 & 60.6 & 41.8 & 41.1 & 41.9 & 48.8 & 50.8 & 52.0 \\
        \hdashline
        \multirow{3}{*}{\textit{\OursFLNoSpace}} & $F_1$ & \textbf{61.9} & \textbf{62.4} & \textbf{62.2} & \textbf{77.2} & \textbf{78.2} & \textbf{77.6} & \textbf{19.2} & \textbf{20.2} & \textbf{21.1} \\
        & $Acc$ & \textbf{76.6} & \textbf{76.7} & \textbf{77.0} & \textbf{77.3} & \textbf{78.2} & \textbf{77.8} & 53.7 & 55.0 & 56.0   \\
        & $MCC$ & \textbf{75.8} & \textbf{75.9} & \textbf{76.2} & \textbf{55.0} & \textbf{56.5} & \textbf{56.2} & 49.7 & 51.1 & 52.2 \\
    \end{tabular}
    \caption{Achieved validation accuracy and F1 score after 100 global aggregation rounds. For all runs $E=1$. $\sigma$ is not reported since for all runs $\sigma < 13.0$ and in most cases $\sigma < 3.0$.}
    \label{tab:metrics100ep_2}
\end{table*}

\begin{table*}[!htb]
    \centering
    \begin{tabular}{l c | c c c c c c c c c }
        \multicolumn{2}{l}{Algorithm} & \multicolumn{3}{c}{FEMNIST} & \multicolumn{3}{c}{CelebA} & \multicolumn{3}{c}{Shakespeare}  \\
        & [\%] / E & 1 & 2 & 4 & 1 & 2 & 4 & 1 & 2 & 4 \\
        \hline
        \multirow{3}{*}{FedAvg} & $F_1$ & 33.1 & 57.7 & 64.4 & 70.5 & 77.0 & 83.8 & 17.4 & 19.4 & \textbf{18.3}   \\
        & $Acc$ & 59.3 & 73.4 & 76.6 & 70.6 & 77.2 & \textbf{84.0} & 52.9 & 53.6 & 51.0   \\
        & $MCC$ & 57.9 & 72.5 & 75.8 & 41.6 & 54.9 & 68.4 & 48.9 & 49.7 & 47.0 \\
        \hdashline
        \multirow{3}{*}{FedAdam} & $F_1$ & 48.1 & 53.1 & 58.1 & 34.2 & 34.0 & 34.0 & 18.1 & 19.0 & 17.9   \\
        & $Acc$ & 66.6 & 69.2 & 71.6 & 51.6 & 51.6 & 51.6 & \textbf{53.8} & \textbf{54.2} & \textbf{52.0}   \\
        & $MCC$ & 65.4 & 68.2 & 70.6 & 1.0 & 0.0 & 0.0 & 49.8 & \textbf{50.3} & \textbf{48.0}\\
        \hdashline
        \multirow{3}{*}{FedAwS} & $F_1$ & 54.8 & 63.8 & 54.8 & 72.2 & 77.9 & 83.2 & 18.3 & 19.6 & 18.4   \\
        & $Acc$ & 73.1 & 77.1 & 70.0 & 72.3 & 78.1 & 83.3 & 53.3 & 53.8 & 51.1   \\
        & $MCC$ & 72.2 & 76.3 & 69.0 & 44.9 & 56.6 & 66.8 & 49.3 & 49.9 & 47.1  \\
        \hdashline
        \multirow{3}{*}{FedProx} & $F_1$ & 32.8 & 57.7 & 60.3 & 70.4 & 77.0 & \textbf{83.9} & 12.5 & 13.5 & 13.9   \\
        & $Acc$ & 59.2 & 73.4 & 74.1 & 70.6 & 77.2 & \textbf{84.0} & 46.0 & 47.6 & 47.6   \\
        & $MCC$ & 57.7 & 72.5 & 73.2 & 41.4 & 55.0 & \textbf{68.5} & 41.2 & 43.0 & 43.2 \\
        \hdashline
        \multirow{3}{*}{FedAMS} & $F_1$ & 46.8 & 53.2 & 57.8 & 36.3 & 34.1 & 34.1 & 18.2 & 19.0 & 17.8  \\
        & $Acc$ & 65.8 & 69.2 & 71.4 & 52.6 & 51.6 & 51.6 & 54.0 & 54.1 & \textbf{52.0}  \\
        & $MCC$ & 64.6 & 68.2 & 70.5 & 4.7 & 0.0 & 0.0 & \textbf{50.0} & 50.2 & \textbf{48.0} \\
        \hdashline
        \multirow{3}{*}{MOON} & $F_1$ & 33.4 & 57.9 & 64.6 & 70.7 & 76.6 & 83.2 & 17.6 & 18.0 & 17.2   \\
        & $Acc$ & 58.2 & 73.3 & 76.7 & 70.9 & 76.9 & 83.3 & 52.8 & 53.2 & 50.0   \\
        & $MCC$ & 56.8 & 72.4 & 75.9 & 41.8 & 54.6 & 67.2 & 48.8 & 49.3 & 45.9 \\
        \hdashline
        \multirow{3}{*}{\textit{\OursFLNoSpace}} & $F_1$ & \textbf{61.9} & \textbf{63.3} & \textbf{60.8} & \textbf{77.2} & \textbf{79.9} & 81.4 & \textbf{19.2} & \textbf{19.5} & 18.1 \\
        & $Acc$ & \textbf{76.6} & \textbf{76.2} & \textbf{74.2} & \textbf{77.3} & \textbf{80.0} & 81.5 & 53.7 & 53.8 & 51.0   \\
        & $MCC$ & \textbf{75.8} & \textbf{75.4} & \textbf{73.4} & \textbf{55.0} & \textbf{60.2} & 63.2 & 49.7 & 50.0 & 47.0 \\
    \end{tabular}
    \caption{Achieved validation accuracy and F1 score after 100 global aggregation rounds (or before overfitting and model crash). For all runs $C=8$. $\sigma$ is not reported since for all runs $\sigma < 13.0$ and in most cases $\sigma < 3.0$.}
    \label{tab:metrics100ep_3}
\end{table*}

\hspace{1cm}
\subsection{\centering Kernelization}
We extended the vanilla~\OursFL with kernelization during SVM fitting and investigated the influence of kernel on model performance. Specifically, we benchmarked polynomial kernel $\mathcal{K}^{poly}(x_1, x_2) = (a x_1^\tau \cdot x_2 + b)^d$ of different degrees ($d = 2, 3, 4$), rbf kernel $\mathcal{K}^{rbf}(x_1, x_2) = \text{exp}(-a ||x_1 - x_2||^2)$, and sigmoid kernel $\mathcal{K}^{sig}(x_1, x_2) = \text{tanh}(a x_1^\tau \cdot x_2 + b)$ on the CelebA dataset with all coefficients $a$ and biases $b$ set to 1.0. The experiments were run with a single random seed for 200 global aggregation rounds. The number of participating clients $C$ in each aggregation round was 8, and each of them trained its local model for $E=1$ epoch. The obtained results are given in Table~\ref{tab:ablation_kernel}. 
\begin{table}[!ht]
    \centering
    \begin{tabular}{l|c|c c|c c| c c | c | c}
          kernel & linear & \multicolumn{2}{c|}{poly (d=2)} & \multicolumn{2}{c|}{poly (d=3)} & \multicolumn{2}{c|}{poly (d=4)} & rbf & sig \\
          server lr & $ 1e^{-2} $ & $1e^{-4}$ & $1e^{-2}$ & $1e^{-4}$ & $1e^{-2}$ & $1e^{-4}$ & $1e^{-1}$ & - & - \\
          \hline
         $F_1$[\%] & 82.1 & 78.4 & 82.0 & 78.3 & 82.8 & 78.4 & 85.2 & 76.8 & 77.3\\ 
          BCE loss & 0.44 & 0.46 & 0.92 & 0.46 & 1.12 & 0.46 & 2.43 & 0.51 & 0.50\\ 
    \end{tabular}
    \caption{Influence of kernel functions.}
    \label{tab:ablation_kernel}
\end{table}
For polynomial kernel, the degree and server learning rate have a large impact: a higher degree and higher server learning rate generally lead to better F1 but also overfitting, mostly due to the simplicity of the task (binary) and the scarcity of participating clients (only 8). In comparison, the rbf kernel and sigmoid kernel are not sensitive to server learning rate when coefficient and bias are 1, and both kernels do not yield improvement. Our results show that kernelized SVM can be used for TurboSVM-FL, and we believe with kernelization, more complex FL tasks can be addressed.

\end{document}